\documentclass[runningheads]{llncs}

\usepackage{microtype}
\usepackage{graphicx}
\usepackage{subfigure}
\usepackage{booktabs} 

\usepackage[breaklinks]{hyperref}

\usepackage{url}
\usepackage{breakurl}

\usepackage{threeparttable}
\usepackage{diagbox}

\usepackage{amsmath,amssymb,mathtools}
\usepackage{mathrsfs}
\usepackage{algorithm,algorithmic}
\usepackage{xspace}
\usepackage{bm}
\usepackage{stfloats}
\usepackage{tablefootnote}
\usepackage{enumerate}
\usepackage{multirow}
\usepackage{xcolor}

\newcommand{\momentum}{\textsc{Momentum}}
\newcommand{\adagrad}{\textsc{Adagrad}}

\newcommand{\adam}{\textsc{Adam}}
\newcommand{\amsgrad}{\textsc{AMSGrad}}
\newcommand{\adahessian}{\textsc{AdaHessian}}
\newcommand{\ftrl}{\textsc{FTRL}}
\newcommand{\diag}{\text{diag}\xspace}
\newcommand{\tabincell}[2]{\begin{tabular}{@{}#1@{}}#2\end{tabular}}

\DeclareMathOperator*{\argmin}{arg\,min}
\DeclareMathOperator{\sign}{sign}

\begin{document}

\title{Adaptive Optimizers with Sparse Group Lasso for Neural Networks in CTR Prediction}
\titlerunning{Adaptive Optimizers with Sparse Group Lasso}

 \author{
   Yun Yue, Yongchao Liu, Suo Tong,  Minghao Li, Zhen Zhang, Chunyang Wen, Huanjun Bao, Lihong Gu, Jinjie Gu, Yixiang Mu
}
\authorrunning{Yun Yue et al.}
\institute{Ant Group, No. 556 Xixi Road, Xihu district, Hangzhou, Zhejiang Province, China
\email{\{yueyun.yy, yongchao.ly, tongsuo.ts, chris.lmh, elliott.zz, chengfu.wcy, alex.bao, lihong.glh, jinjie.gujj, yixiang.myx\}@antgroup.com}}
%
\toctitle{Adaptive Optimizers with Sparse Group Lasso for Neural Networks in CTR Prediction}
\tocauthor{Yun Yue, Yongchao Liu, Suo Tong,  Minghao Li, Zhen Zhang, Chunyang Wen, Huanjun Bao, Lihong Gu, Jinjie Gu, Yixiang Mu}
\maketitle              

\begin{abstract}
	We develop a novel framework that adds the regularizers of the sparse group lasso to a family of adaptive optimizers in deep learning, such as \momentum, \adagrad, \adam, \amsgrad, \adahessian{}, and create a new class of optimizers, which are named \textsc{Group Momentum}, \textsc{Group Adagrad}, \textsc{Group Adam}, \textsc{Group AMSGrad} and \textsc{Group AdaHessian}, etc., accordingly. We establish theoretically proven convergence guarantees in the stochastic convex settings, based on primal-dual methods. We evaluate the regularized effect of our new optimizers on three large-scale real-world ad click datasets with state-of-the-art deep learning models. The experimental results reveal that compared with the original optimizers with the post-processing procedure which uses the magnitude pruning method, the performance of the models can be significantly improved on the same sparsity level. Furthermore, in comparison to the cases without magnitude pruning, our methods can achieve extremely high sparsity with significantly better or highly competitive performance. The code is available at this link\footnote{https://github.com/intelligent-machine-learning/tfplus/tree/main/tfplus}.
	
\end{abstract}

\section{Introduction}
\label{sec:1}

With the development of deep learning, deep neural network (DNN) models have been widely used in various machine learning scenarios such as search, recommendation and advertisement, and achieved significant improvements. In the last decades, different kinds of optimization methods based on the variations of stochastic gradient descent (SGD) have been invented for training DNN models. However, most optimizers cannot directly produce sparsity which has been proven effective and efficient for saving computational resource and improving model performance especially in the scenarios of very high-dimensional data. Meanwhile, the simple rounding approach is very unreliable due to the inherent low accuracy of these optimizers.

In this paper, we develop a new class of optimization methods, that adds the regularizers especially sparse group lasso to prevalent adaptive optimizers, and retains the characteristics of the respective optimizers. Compared with the original optimizers with the post-processing procedure which use the magnitude pruning method, the performance of the models can be significantly improved on the same sparsity level. Furthermore, in comparison to the cases without magnitude pruning, the new optimizers can achieve extremely high sparsity with significantly better or highly competitive performance. In this section, we describe the two types of optimization methods, and explain the motivation of our work.

\subsection{Adaptive Optimization Methods}

Due to the simplicity and effectiveness, adaptive optimization methods \cite{sgd,momentum2,adagrad,adadelta,adam,amsgrad,adahessian} have become the de-facto standard algorithms used in deep learning. There are multiple variants, but they can be represented using the general update formula \cite{amsgrad}:
\begin{equation}
x_{t+1} = x_t-\alpha_{t}m_t/\sqrt{V_t},
\label{eq:general-formula}
\end{equation}
where $\alpha_{t}$ is the step size, $m_t$ is the first moment term which is the weighted average of gradient $g_t$ and $V_t$ is the so called second moment term that adjusts updated velocity of variable $x_t$ in each direction. Here, $\sqrt{V_t}:=V_t^{1/2}$, $m_t/\sqrt{V_t}:=\sqrt{V_t}^{-1} \cdot m_t$. By setting different $m_{t}$, $V_{t}$ and $\alpha_{t}$ , we can derive different adaptive optimizers including $\momentum$ \cite{momentum2}, $\adagrad$ \cite{adagrad}, $\adam$ \cite{adam}, $\amsgrad$ \cite{amsgrad} and $\adahessian$ \cite{adahessian}, etc. See Table~\ref{tab:general-formula}.
\begin{table*}[!hbtp]
	\small
	\caption{
		Adaptive optimizers with choosing different $m_t$, $V_t$ and $\alpha_{t}$.
	}
	\centering
	\label{tab:general-formula}
	\begin{threeparttable}
		\centering
					\begin{tabular}{cccc}
						\toprule
						\textbf{Optimizer} & \bm{$m_t$} & \bm{$V_t$} & \bm{$\alpha_{t}$} \\
						\midrule
						\textsc{Sgd} & $g_t$ & $\mathbb{I}$ & $\frac{\alpha}{\sqrt{t}}$ \\
						\textsc{Momentum} & $\gamma{}m_{t-1} + g_t$ & $\mathbb{I}$ & $\alpha$ \\
						\textsc{Adagrad} & $g_t$ & $\diag(\sum_{i=1}^{t} g_i^2) / t$ & $\frac{\alpha}{\sqrt{t}}$ \\
						\textsc{Adam} & $\beta_{1} m_{t - 1} + (1 - \beta_{1}) g_t$ & $\beta_{2} V_{t-1} + (1 - \beta_{2}) \diag(g_t^2)$ & $\frac{\alpha{}\sqrt{1 - \beta_{2}^{t}}}{1 - \beta_{1}^{t}}$ \\
						\textsc{AMSGrad} & $\beta_{1} m_{t - 1} + (1 - \beta_{1}) g_t$ & $\max(V_{t-1}, \beta_{2} V_{t-1} + (1 - \beta_{2}) \diag(g_t^2))$ & $\frac{\alpha{}\sqrt{1 - \beta_{2}^{t}}}{1 - \beta_{1}^{t}}$ \\
						\textsc{AdaHessian} & $\beta_{1} m_{t - 1} + (1 - \beta_{1}) g_t$ & $\beta_{2} V_{t-1} + (1 - \beta_{2}) D_t^2$ \tnote{*} & $\frac{\alpha{}\sqrt{1 - \beta_{2}^{t}}}{1 - \beta_{1}^{t}}$ \\
						\bottomrule
					\end{tabular}
				\begin{tablenotes}
					\item[*] $D_t = \diag(H_t)$, where $H_t$ is the Hessian matrix. 
				\end{tablenotes}
	\end{threeparttable}
	\vskip -0.15in
\end{table*}

\subsection{Regularized Optimization Methods}

Follow-the-regularized-leader (\ftrl) \cite{ftrl2,adclickprediction} has been widely used in click-through rates (CTR) prediction problems, which adds $\ell_{1}$-regularization (lasso) to logistic regression and can effectively balance the performance of the model and the sparsity of features. The update formula \cite{adclickprediction} is: 
\begin{equation}
x_{t+1} = \argmin_{x} g_{1:t} \cdot x + \frac{1}{2} \sum_{s=1}^{t} \sigma{}_{s} \|x - x_s\|_2^2 + \lambda_{1}\|x\|_1,
\label{eq:ftrl}
\end{equation}
where $g_{1:t} = \sum{}_{s=1}^{t} g_s$, $\frac{1}{2} \sum_{s=1}^{t} \sigma{}_{s} \|x - x_s\|_2^2$ is the strong convex term that stabilizes the algorithm and $\lambda_{1}\|x\|_1$ is the regularization term that produces sparsity. However, it doesn't work well in DNN models since one input feature can correspond to multiple weights and lasso only can make single weight zero hence can't effectively delete features.

To solve above problem, \cite{groupftrl} adds the $\ell{}_{21}$-regularization (group lasso) to \ftrl{}, which is named G-FTRL. \cite{onlinegrouplasso} conducts the research on a group lasso method for online learning that adds $\ell{}_{21}$-regularization to the algorithm of Dual Averaging (DA) \cite{primaldual}, which is named DA-GL. Even so, these two methods cannot be applied to other optimizers. Different scenarios are suitable for different optimizers in the deep learning fields. For example, \momentum{} \cite{momentum2} is typically used in computer vision; \adam{} \cite{adam} is used for training transformer models for natural language processing; and \adagrad{} \cite{adagrad} is used for recommendation systems. If we want to produce sparsity of the model in some scenario, we have to change optimizer which probably influence the performance of the model.

\subsection{Motivation}

Eq.~\eqref{eq:general-formula} can be rewritten into this form:
\begin{equation}
x_{t+1} = \argmin_{x} m_t\cdot x + \frac{1}{2\alpha_{t}}\|\sqrt{V_t}^\frac{1}{2}(x-x_t)\|_2^2.
\label{eq:general-formula2}
\end{equation}
Furthermore, we can rewrite Eq.~\eqref{eq:general-formula2} into
\begin{equation}
x_{t+1} = \argmin_{x} m_{1:t} \cdot x +
\sum_{s=1}^{t}\frac{1}{2\alpha_{s}}\|Q_s^\frac{1}{2}(x-x_s)\|_2^2,
\label{eq:general-formula3}
\end{equation}
where $m_{1:t}=\sum_{s=1}^{t} m_{s}$, $\sum_{s=1}^{t} Q_{s}/\alpha_{s}=\sqrt{V_t}/\alpha_{t}$. It is easy to prove that Eq.~\eqref{eq:general-formula2} and Eq.~\eqref{eq:general-formula3} are equivalent using the method of induction. The matrices $Q_s$ can be interpreted as generalized learning rates. To our best knowledge, $V_t$ of Eq.~\eqref{eq:general-formula} of all the adaptive optimization methods are diagonal for the computation simplicity. Therefore, we consider $Q_s$ as diagonal matrices throughout this paper.

We find that Eq.~\eqref{eq:general-formula3} is similar to Eq.~\eqref{eq:ftrl} except for the regularization term. Therefore, we add the regularization term $\Psi{}(x)$ to Eq.~\eqref{eq:general-formula3}, which is the sparse group lasso penalty also including $\ell_{2}$-regularization that can diffuse weights of neural networks. The concrete formula is:
\begin{equation}
\Psi_t(x) = \sum_{g=1}^{G} \Big(\lambda_{1}\|x^g\|_{1} + \lambda_{21} \sqrt{d_{x^g}}\|A_{t}^{\frac{1}{2}} x^g\|_{2} \Big) + \lambda_{2}\|x\|_{2}^{2},
\label{eq:sgl}
\end{equation}
where $\lambda_{1}$, $\lambda_{21}$, $\lambda_{2}$ are regularization parameters of $\ell{}_{1}$, $\ell{}_{21}$, $\ell{}_{2}$ respectively, $G$ is the total number of groups of weights, $x^g$ is the weights of group $g$ and $d_{x^g}$ is the size of group $g$. In DNN models, each group is defined as the set of outgoing weights from a unit which can be an input feature, or a hidden neuron, or a bias unit (see, e.g., \cite{groupsparse}). $A_t$ can be arbitrary positive matrix satisfying $A_{t+1} \succeq A_t$, e.g., $A_t = \mathbb{I}$. In Section~\ref{subsec:2.1}, we let $A_t = (\sum_{s=1}^{t}\frac{Q_s^g}{2\alpha_{s}} + \lambda_{2}\mathbb{I})$ just for solving the closed-form solution directly, where $Q_s^g$ is a diagonal matrix whose diagonal elements are part of $Q_s$ corresponding to $x_g$. The ultimate update formula is:
\begin{equation}
x_{t+1} = \argmin_{x} m_{1:t} \cdot x + \sum_{s=1}^{t}\frac{1}{2\alpha_{s}}\|Q_s^\frac{1}{2}(x-x_s)\|_2^2 + \Psi_t(x).
\label{eq:ada-sgl}
\end{equation}

\subsection{Outline of Contents}

The rest of the paper is organized as follows. In Section~\ref{subsec:1.5}, we introduce the necessary notations and technical background.

In Section~\ref{sec:2}, we present the closed-form solution of Eq.~\eqref{eq:general-formula3} and the algorithm of general framework of adaptive optimization methods with sparse group lasso. We prove the algorithm is equivalent to adaptive optimization methods when regularization terms vanish. In the end, we give two concrete examples of the algorithm.

In Section~\ref{sec:3}, we derive the regret bounds of the method and convergence rates.

In Section~\ref{sec:4}, we validate the performance of new optimizers in the public datasets. 

In Section~\ref{sec:5}, we summarize the conclusion.

 Appendices~\ref{appendix:ada-group-lasso}-\ref{appendix:add_prof} contain technical proofs of our main results and Appendix~\ref{appendix:emp_rst} includes the additional details of the experiments of Section~\ref{sec:4}.

\subsection{Notations and Technical Background}
\label{subsec:1.5}

We use lowercase letters to denote scalars and vectors, and uppercase letters to denote matrices. We denote a sequence of vectors by subscripts, that is, $x_1,\dots,x_t$, and entries of each vector by an additional subscript, e.g., $x_{t,i}$. We use the notation $g_{1:t}$ as a shorthand for $\sum_{s=1}^{t} g_s$. Similarly we write $m_{1:t}$ for a sum of the first moment $m_t$, and $f_{1:t}$ to denote the function $f_{1:t}(x) = \sum_{s=1}^{t} f_s(x)$. Let $M_t = [m_1 \cdots m_t]$ denote the matrix obtained by concatenating the vector sequence $\{m_t\}_{t\geq 1}$ and $M_{t,i}$ denote the $i$-th row of this matrix which amounts to the concatenation of the $i$-th component of each vector. The notation $A \succeq 0$ (resp. $A \succ 0$) for a matrix A means that A is symmetric and positive semidefinite (resp. definite). Similarly, the notations $A \succeq B$ and $A \succ B$ mean that $A - B \succeq 0$ and $A - B \succ 0$ respectively, and both tacitly assume that $A$ and $B$ are symmetric. Given $A \succeq 0$, we write $A^{\frac{1}{2}}$ for the square root of $A$, the unique $X \succeq 0$ such that $XX = A$ (\cite{ftrl2}, Section 1.4).

Let $\mathcal{E}$ be a finite-dimension real vector space, endowed with the Mahalanobis norm $\|\cdot\|_A$ which is denoted by $\|\cdot\|_A = \sqrt{\left<\cdot, A\cdot\right>}$ as induced by $A \succ 0$.
Let $\mathcal{E}^{*}$ be the vector space of all linear functions on $\mathcal{E}$. The dual space $\mathcal{E}^{*}$ is endowed with the dual norm $\|\cdot\|^{*}_A = \sqrt{\left< \cdot, A^{-1}\cdot\right>}$. 

Let $\mathcal{Q}$ be a closed convex set in $\mathcal{E}$. A continuous function $h(x)$ is called \textit{strongly convex} on $\mathcal{Q}$ with norm $\|\cdot\|_{H}$ if $\mathcal{Q} \subseteq \text{dom}\ h$ and there exists a constant $\sigma > 0$ such that for all $x, y \in \mathcal{Q}$ and $\alpha \in [0, 1]$ we have
\begin{equation*}
h(\alpha x + (1 - \alpha)y) \leq \alpha h(x) + (1 - \alpha)h(y) - \frac{1}{2}\sigma\alpha(1 - \alpha)\|x - y\|^2_H.
\end{equation*}
The constant $\sigma$ is called the \textit{convexity parameter} of $h(x)$, or the \textit{modulus} of strong convexity. We also denote by $\|\cdot\|_h = \|\cdot\|_H$. Further, if $h$ is differentiable, we have 
\begin{equation*}
	h(y) \geq h(x) + \left<\nabla{}h(x), y - x\right> + \frac{\sigma}{2} \| x - y\|^2_h.
\end{equation*}

We use online convex optimization as our analysis framework. On each round $t = 1, \dots, T$, a convex loss function $f_t: \mathcal{Q} \mapsto \mathbb{R}$ is chosen, and we pick a point $x_t \in \mathcal{Q}$ hence get loss $f_t(x_t)$. Our goal is minimizing the \textit{regret} which is defined as the quantity
\begin{equation}
	\mathcal{R}_T = \sum_{t=1}^{T} f_t(x_t) - \min_{x \in \mathcal{Q}}\sum_{t=1}^{T} f_t(x). 
	\label{eq:regret}
\end{equation}
Online convex optimization can be seen as a generalization of stochastic convex optimization. Any regret minimizing algorithm can be converted to a stochastic optimization algorithm with convergence rate $O(\mathcal{R}_{T}/T)$ using an online-to-batch conversion technique \cite{online_to_batch}.

In this paper, we assume $\mathcal{Q}\equiv\mathcal{E}=\mathbb{R}^n$, hence we have $\mathcal{E}^*=\mathbb{R}^n$.
 We write $s^{T}x$ or $s\cdot x$ for the standard inner product between $s, x \in \mathbb{R}^n$. For the standard Euclidean norm, $\|x\| = \|x\|_2 = \sqrt{\left<x, x\right>}$ and $\|s\|_{*} = \|s\|_2$. We also use $\|x\|_1 = \sum_{i=1}^{n} |x^{(i)}|$ and $\|x\|_{\infty} = \max{}_i|x^{(i)}|$ to denote $\ell_{1}$-norm and $\ell_{\infty}$-norm respectively, where $x^{(i)}$ is the $i$-th element of $x$.

\section{Algorithm}
\label{sec:2}
\subsection{Closed-form  Solution}
\label{subsec:2.1}

We will derive the closed-form solution of Eq.~\eqref{eq:ada-sgl} with specific $A_t$ and Algorithm~\ref{alg:ada-group-lasso} with slight modification in this section. We have the following theorem.

\begin{theorem}
\label{thm:ada-group-lasso}
Given $A_t = (\sum_{s=1}^{t}\frac{Q_s^g}{2\alpha_{s}} + \lambda_{2}\mathbb{I})$ of Eq.~\eqref{eq:sgl}, $z_t = z_{t-1} + m_t - \frac{Q_t}{\alpha_t}x_t$ at each iteration $t = 1, \dots, T$ and $z_0 = \mathbf{0}$, the optimal solution of Eq.~\eqref{eq:ada-sgl} is updated accordingly as follows:
\begin{equation}
\begin{aligned}
x_{t+1} = (\sum_{s=1}^{t} \frac{Q_s}{\alpha_{s}} + 2\lambda_{2}\mathbb{I})^{-1}\max(1-\frac{\sqrt{d_{x_t^g}}\lambda_{21}}{\|\tilde{s}_t\|_2}, 0)s_t
\label{eq:ada-group-lasso}
\end{aligned}
\end{equation}
where the $i$-th element of $s_t$ is defined as 
\begin{equation}
s_{t,i} = \left\{ 
\begin{array}{ll}
0 & \textrm{if $|z_{t,i}| \le \lambda_1$,}\\
\sign(z_{t,i})\lambda_1 - z_{t,i} & \textrm{otherwise,}
\end{array} \right.
\label{eq:s_t}
\end{equation}
$\tilde{s}_t$ is defined as 
\begin{equation}
	\tilde{s}_t = (\sum_{s=1}^{t} \frac{Q_s}{2\alpha_{s}} + \lambda_{2}\mathbb{I})^{-1} s_t
	\label{eq:st_2}
\end{equation}
and $\sum_{s=1}^{t} \frac{Q_s}{\alpha_{s}}$ is the diagonal and positive definite matrix.
\end{theorem}

The proof of Theorem~\ref{thm:ada-group-lasso} is given in Appendix~\ref{appendix:ada-group-lasso}. Here $\tilde{s}_t$ can be considered as the weighted average of $s_t$. We slightly modify \eqref{eq:ada-group-lasso} where we replace $\tilde{s}_t$ with $s_t$ in practical algorithms. Our purpose is that the $\ell_{21}$-regularization
does not depend on the second moment terms and other hyperparameters such as $\alpha_{s}$ and $\lambda_{2}$. The empirical experiment will also show that the algorithm using $s_t$ can improve accuracy over using $\tilde{s}_t$ in the same level of sparsity in Section~\ref{subsec:4.4}. Therefore, we get Algorithm~\ref{alg:ada-group-lasso}. Furthermore, we have the following theorem which shows the relationship between Algorithm~\ref{alg:ada-group-lasso} and adaptive optimization methods. The proof is given in Appendix~\ref{appendix:equivalent}.

\begin{algorithm}[htbp]\small
	\caption{Generic framework of adaptive optimization methods with sparse group lasso}
	\label{alg:ada-group-lasso}
	\begin{algorithmic}[1]
		\STATE {\bfseries Input:} parameters $\lambda_{1}, \lambda_{21}, \lambda_{2}$\\
		$x_1 \in \mathbb{R}^n$, step size $\{\alpha_t > 0\}_{t=0}^T$, sequence of functions $\{\phi_t, \psi_t\}_{t=1}^T$, initialize $z_0=\mathbf{0}, V_0=\mathbf{0}$
		\FOR{$t=1$ {\bfseries to} $T$}
		\STATE $g_t = \nabla f_t(x_t)$
		\STATE $m_t = \phi_t(g_1, \dots, g_t)$ and $V_t = \psi_t(g_1 , \dots, g_t)$
		\STATE $\frac{Q_t}{\alpha_{t}} = \frac{\sqrt{V_t}}{\alpha_{t}} - \frac{\sqrt{V_{t-1}}}{\alpha_{t-1}}$
		\STATE $z_t\leftarrow z_{t-1} + m_t - \frac{Q_t}{\alpha_{t}} x_t$
		\FOR{$i \in \{1, \dots, n\}$}
		\STATE $s_{t,i} = \left\{ 
		\begin{array}{ll}
		0 & \textrm{if $|z_{t,i}| \le \lambda_1$}\\
		\sign(z_{t,i})\lambda_1 - z_{t,i} & \textrm{otherwise.}
		\end{array} \right.$
		\ENDFOR
		\STATE $x_{t+1} = (\frac{\sqrt{V_t}}{\alpha_{t}} + 2\lambda_{2}\mathbb{I})^{-1}\max(1-\frac{\sqrt{d_{x_t^g}}\lambda_{21}}{\|s_t\|_2}, 0)s_t$
		\ENDFOR
	\end{algorithmic}
\end{algorithm}


\begin{theorem}
	\label{thm:equivalent}
	If regularization terms of Algorithm~\ref{alg:ada-group-lasso} vanish, Algorithm~\ref{alg:ada-group-lasso} is equivalent to Eq. \eqref{eq:general-formula}.
\end{theorem}

\subsection{Concrete Examples}
Using Algorithm~\ref{alg:ada-group-lasso}, we can easily derive the new optimizers based on \adam{ }\cite{adam}, \adagrad{ }\cite{adagrad} which we call \textsc{Group Adam}, \textsc{Group Adagrad} respectively.
\subsubsection*{\textsc{Group Adam}}

The detail of the algorithm is given in Algorithm~\ref{alg:group-adam}. From Theorem~\ref{thm:equivalent}, we know that when $\lambda_{1}$, $\lambda_{2}$, $\lambda_{21}$ are all zeros, Algorithm~\ref{alg:group-adam} is equivalent to \adam{ }\cite{adam}.

\begin{figure}[!tb]
	\begin{minipage}[t]{0.49\columnwidth}
		\centering
		\begin{algorithm}[H]\small
			\caption{Group Adam}
			\label{alg:group-adam}
			\begin{algorithmic}[1]
				\STATE {\bfseries Input:} parameters $\lambda_{1}, \lambda_{21}, \lambda_{2}$, $\beta_{1}$, $\beta_{2}$, $\epsilon$\\
				$x_1 \in \mathbb{R}^n$, step size $\alpha$, initialize $z_0=\mathbf{0}, \hat{m}_0=\mathbf{0}, \hat{V}_0=\mathbf{0}, V_0=\mathbf{0}$
				\FOR{$t=1$ {\bfseries to} $T$}
				\STATE $g_t = \nabla f_t(x_t)$
				\STATE $\hat{m}_{t}\leftarrow \beta_{1}\hat{m}_{t-1} + (1 - \beta_{1})g_{t}$
				\STATE $m_t = \hat{m}_t / (1 - \beta_{1}^{t})$
				\STATE $\hat{V}_t\leftarrow \beta_{2}\hat{V}_{t-1} + (1 - \beta_{2})\diag(g_t^2)$
				\STATE $V_t = \hat{V}_t / (1 - \beta_{2}^{t})$
				\STATE $\epsilon \leftarrow \epsilon / \sqrt{1 - \beta_{2}^t}$
				\STATE $Q_t = \left\{
				\begin{array}{ll}
				\sqrt{V_t} - \sqrt{V_{t-1}} + \epsilon\mathbb{I} & t = 1\\
				\sqrt{V_t} - \sqrt{V_{t-1}} & t > 1
				\end{array} \right.$
				\STATE $z_t\leftarrow z_{t-1} + m_t - \frac{1}{\alpha}Q_t x_t$
				\FOR{$i \in \{1, \dots, n\}$}
				\STATE $s_{t,i} = -\sign(z_{t,i})\max(|z_{t,i}| - \lambda_{1}, 0)$
				\ENDFOR
				\STATE $x_{t+1} = (\frac{\sqrt{V_t} + \epsilon\mathbb{I}}{\alpha} + 2\lambda_{2}\mathbb{I})^{-1}\max(1-\frac{\sqrt{d_{x_t^g}}\lambda_{21}}{\|s_t\|_2}, 0)s_t$
				\ENDFOR
			\end{algorithmic}
		\end{algorithm}
	\end{minipage}
	\hskip 0.1in
	\begin{minipage}[t]{0.49\columnwidth}
		\begin{algorithm}[H]\small
			\caption{Group Adagrad}
			\label{alg:group-adagrad}
			\begin{algorithmic}[1]
				\STATE {\bfseries Input:} parameters $\lambda_{1}, \lambda_{21}, \lambda_{2}, \epsilon$\\
				$x_1 \in \mathbb{R}^n$, step size $\alpha$, initialize $z_0=\mathbf{0}, V_0=\mathbf{0}$
				\FOR{$t=1$ {\bfseries to} $T$}
				\STATE $g_t = \nabla f_t(x_t)$
				\STATE $m_t = g_t$
				\STATE $V_t = \left\{
				\begin{array}{ll}
				V_{t-1} + \diag(g_t^2) + \epsilon\mathbb{I} & t = 1\\
				V_{t-1} + \diag(g_t^2) & t > 1
				\end{array}
				\right.$
				\STATE $Q_t =\sqrt{V_t} - \sqrt{V_{t-1}}$
				\STATE $z_t\leftarrow z_{t-1} + m_t - \frac{1}{\alpha}Q_t x_t$
				\FOR{$i \in \{1, \dots, n\}$}
				\STATE $s_{t,i} = -\sign(z_{t,i})\max(|z_{t,i}| - \lambda_{1}, 0)$
				\ENDFOR
				\STATE $x_{t+1} = (\frac{\sqrt{V_t}}{\alpha} + 2\lambda_{2}\mathbb{I})^{-1}\max(1-\frac{\sqrt{d_{x_t^g}}\lambda_{21}}{\|s_t\|_2}, 0)s_t$
				\ENDFOR
			\end{algorithmic}
		\end{algorithm}
	\end{minipage}
\end{figure}

\subsubsection*{\textsc{Group Adagrad}}

The detail of the algorithm is given in Algorithm~\ref{alg:group-adagrad}. Similarly, from Theorem~\ref{thm:equivalent}, when $\lambda_{1}$, $\lambda_{2}$, $\lambda_{21}$ are all zeros, Algorithm~\ref{alg:group-adagrad} is equivalent to \adagrad{ }\cite{adagrad}. Furthermore, we can find that when $\lambda_{21}=0$, Algorithm~\ref{alg:group-adagrad} is equivalent to \ftrl{ }\cite{adclickprediction}. Therefore, \textsc{Group Adagrad} can also be called \textsc{Group FTRL} from the research of \cite{groupftrl}.

Similarly, \textsc{Group Momentum}, \textsc{Group AMSGrad}, \textsc{Group AdaHessian}, etc., can be derived from \momentum{} \cite{momentum2}, \amsgrad{} \cite{amsgrad}, \textsc{AdaHessian} \cite{adahessian}, etc., with the same framework and we will not list the details.

\section{Convergence and Regret Analysis}
\label{sec:3}

Using the framework developed in \cite{primaldual,dualavg,adagrad}, we have the following theorem providing the bound of the regret.
\begin{theorem}
	Let the sequence $\{x_t\}$ be defined by the update \eqref{eq:ada-sgl} and 
	\begin{equation}
		x_1 = \argmin_{x\in\mathcal{Q}} \frac{1}{2} \|x-c\|_2^2,
		\label{eq:x1}
	\end{equation}
	where $c$ is an arbitrary constant vector.
	Suppose $f_t(x)$ is convex for any $t \geq 1$ and there exists an optimal solution $x^*$ of $\sum_{t=1}^{T} f_t(x)$, i.e., $x^* = \argmin_{x\in\mathcal{Q}} \sum_{t=1}^{T} f_t(x)$, which satisfies the condition 
	\begin{equation}
		\left<m_{t - 1}, x_t - x^*\right> \geq 0, \quad t \in [T],
		\label{eq:cond}
	\end{equation}
	where $m_t$ is the weighted average of the gradient $f_t(x_t)$ and $[T] = \{1, \dots, T\}$ for simplicity. Without loss of generality, we assume 
	\begin{equation}
	m_t = \gamma m_{t-1} + g_t,
	\label{eq:mt}
	\end{equation}
	where $\gamma < 1$ and $m_0 = 0$.
	Then
	\begin{equation}
		\mathcal{R}_T \leq \Psi_T(x^*) + \sum_{t=1}^{T} \frac{1}{2\alpha_{t}}\|Q_t^{\frac{1}{2}}(x^*-x_t)\|_2^2 + \frac{1}{2}\sum_{t=1}^{T} \|m_t\|_{h_{t-1}^*}^2,
		\label{eq:rt_up}
	\end{equation}
	where $\|\cdot\|_{h_t^*}$ is the dual norm of $\|\cdot\|_{h_t}$. $h_t$ is $1$-strongly convex with respect to $\|\cdot\|_{\sqrt{V_t}/\alpha_{t}}$ for $t\in{}[T]$ and $h_0$ is $1$-strongly convex with respect to $\|\cdot\|_2$.
	\label{thm:convergence}
\end{theorem}

The proof of Theorem~\ref{thm:convergence} is given in Appendix~\ref{appendix:convergence}. Since in most of adaptive optimizers, $V_t$ is the weighted average of $\diag(g_t^2)$, without loss of generality, we assume $\alpha_{t} = \alpha$ and
\begin{equation}
		V_t = \eta V_{t-1} + \diag(g_t^2),\quad t \geq 1,
		\label{eq:vt}
\end{equation}
where $V_0 = 0$ and $\eta\leq 1$. Hence, we have the following lemma whose proof is given in Appendix~\ref{appendix:lemma1}.
\begin{lemma}
	Suppose $V_t$ is the weighted average of the square of the gradient which is defined by \eqref{eq:vt}, $\alpha_{t} = \alpha$, $m_t$ is defined by \eqref{eq:mt} and one of the following conditions:
	\begin{enumerate}
		\item $\eta = 1$,
		\item $\eta < 1$, $\eta\geq\gamma$ and $\kappa V_t \succeq V_{t-1}$ for all $t\geq 1$ where $\kappa < 1$.
	\end{enumerate}
	is satisfied. Then we have 
	\begin{equation}
	\sum_{t=1}^{T}\|m_t\|^2_{(\frac{\sqrt{V_t}}{\alpha_{t}})^{-1}} < \frac{2\alpha}{1-\nu} \sum_{i=1}^{d} \|M_{T, i}\|_2,
	\label{eq:vt_4}
	\end{equation} 
	\label{lem:1}
where $\nu = \max(\gamma, \kappa)$ and $d$ is the dimension of $x_t$.
\end{lemma}

 We can always add $\delta^2\mathbb{I}$ to $V_t$ at each step to ensure $V_t \succ 0$. Therefore, $h_t(x)$ is $1$-strongly convex with respect to $\|\cdot\|_{\sqrt{\delta^2\mathbb{I} + V_t}/\alpha_{t}}$. Let $\delta \geq \max_{t\in[T]} \|g_t\|_{\infty}$, for $t >1$, we have
 \begin{equation}
 	\begin{aligned}
 	\|m_t\|_{h_{t-1}^*}^2 &= \left<m_t, \alpha_{t}(\delta^2\mathbb{I} + V_{t-1})^{-\frac{1}{2}} m_t\right> \leq \left<m_t, \alpha_{t}\left(\diag(g_t^2) + \eta V_{t-1}\right)^{-\frac{1}{2}}m_t\right>\\
 	&=\left<m_t, \alpha_{t} V_t^{-\frac{1}{2}} m_t\right> = \|m_t\|^2_{(\frac{\sqrt{V_t}}{\alpha_{t}})^{-1}}.
 	\label{eq:vt_5}
 	\end{aligned} 	
 \end{equation}
 For $t = 1$, we have 
  \begin{equation}
 \begin{aligned}
 \|m_1\|_{h_{0}^*}^2 &= \left<m_1, \alpha_{1}(\delta^2\mathbb{I} + \mathbb{I})^{-\frac{1}{2}} m_1\right> \leq\left<m_1, \alpha_{1}\left(\diag^{-\frac{1}{2}}(g_1^2)\right)m_1\right> \\
 &=\left<m_1, \alpha_{1} V_1^{-\frac{1}{2}} m_1\right> = \|m_1\|^2_{(\frac{\sqrt{V_1}}{\alpha_{1}})^{-1}}.
 \label{eq:vt_6}
 \end{aligned} 	
 \end{equation}
 
 From \eqref{eq:vt_5}, \eqref{eq:vt_6} and Lemma~\ref{lem:1}, we have
 
\begin{lemma}
	Suppose $V_t$, $m_t$, $\alpha_{t}$, $\nu$, $d$ are defined the same as Lemma~\ref{lem:1}, $\max_{t\in[T]} \|g_t\|_{\infty} \leq \delta$, $\|\cdot\|_{h_{t}^*}^2 = \left<\cdot, \alpha_{t}(\delta^2\mathbb{I} + V_t)^{-\frac{1}{2}} \cdot\right>$ for $t\geq 1$ and $\|\cdot\|_{h_0^*}^2 = \left<\cdot, \alpha_{1} \left((\delta^2 + 1)\mathbb{I}\right)^{-\frac{1}{2}} \cdot \right>$. Then
	\begin{equation}
		\sum_{t=1}^{T}\|m_t\|_{h_{t-1}^*}^2 < \frac{2\alpha}{1-\nu} \sum_{i=1}^{d} \|M_{T, i}\|_2.
	\label{eq:vt_7}
	\end{equation}
	\label{lem:2}
\end{lemma}

Therefore, from Theorem~\ref{thm:convergence} and Lemma~\ref{lem:2}, we have

\begin{corollary}
	Suppose $V_t$, $m_t$, $\alpha_{t}$, $h_t^*$, $\nu$, $d$ are defined the same as Lemma~\ref{lem:2}, there exist constants $G$, $D_1$, $D_2$ such that $\max_{t\in[T]} \|g_t\|_{\infty} \leq G \leq\delta$, $\|x^*\|_{\infty} \leq D_1$ and $\max_{t\in[T]}\|x_t-x^*\|_{\infty} \leq D_2$. Then
	\begin{equation}
	   \mathcal{R}_T < dD_1\left(\lambda_{1} + \lambda_{21}(\frac{\sqrt{T}G}{2\alpha} + \lambda_{2})^{\frac{1}{2}} + \lambda_{2}D_1\right) + dG\left(\frac{D_2^2}{2\alpha} + \frac{\alpha}{(1-\nu)^{2}}\right)\sqrt{T}.
	\end{equation}
	\label{coro:1}
\end{corollary}

The proof of Corollary~\ref{coro:1} is given in \ref{appendix:coro1}. Furthermore, from Corollary~\ref{coro:1}, we have

\begin{corollary}
	Suppose $m_t$ is defined as \eqref{eq:mt}, $\alpha_{t} = \alpha$ and satisfies the condition \eqref{eq:vt_7}. There exist constants $G$, $D_1$, $D_2$ such that  $tG^2\mathbb{I} \succeq V_t$, $\max_{t\in[T]} \|g_t\|_{\infty} \leq G$, $\|x^*\|_{\infty} \leq D_1$ and $\max_{t\in[T]}\|x_t-x^*\|_{\infty} \leq D_2$. Then
	\begin{equation}
	\mathcal{R}_T < dD_1\left(\lambda_{1} + \lambda_{21}(\frac{\sqrt{T}G}{2\alpha} + \lambda_{2})^{\frac{1}{2}} + \lambda_{2}D_1\right) + dG\left(\frac{D_2^2}{2\alpha} + \frac{\alpha}{(1-\nu)^{2}}\right)\sqrt{T}.
	\end{equation}
	\label{coro:2}
\end{corollary}
Therefore, we know that the regret of the update \eqref{eq:ada-sgl} is $O(\sqrt{T})$ and can achieve the convergence rate $O(1/\sqrt{T})$ under the conditions of Corollary~\ref{coro:1} or Corollary~\ref{coro:2}.

\section{Experiments}
\label{sec:4}

\subsection{Experiment Setup}
We test the algorithms on three different large-scale real-world datasets with different neural network structures. These datasets are various display ads logs for the purpose of predicting ads CTR. The details are as follows.
\begin{enumerate}[a)]
	\item The Avazu CTR dataset \cite{avazu} contains approximately 40M samples and 22 categorical features over 10 days. In order to handle categorical data, we use the one-hot-encoding based embedding technique (see, e.g., \cite{dcn}, Section 2.1 or \cite{dlrm}, Section 2.1.1) and get 9.4M features in total. For this dataset, the samples from the first 9 days (containing 8.7M one-hot features) are used for training, while the rest is for testing. Our DNN model follows the basic structure of most deep CTR models. Specifically, the model comprises one embedding layer, which maps each one-hot feature into 16-dimensional embeddings, and four fully connected layers (with output dimension of 64, 32, 16 and 1, respectively) in sequence.
	\item The iPinYou dataset\footnote{We only use the data from season 2 and 3 because of the same data schema.} \cite{ipinyou} is another real-world dataset for ad click logs over 21 days. The dataset contains 16 categorical features\footnote{See \url{https://github.com/Atomu2014/Ads-RecSys-Datasets/} for details.}. After one-hot encoding, we get a dataset containing 19.5M instances with 1033.1K input dimensions. We keep the original train/test splitting scheme, where the training set contains 15.4M samples with 937.7K one-hot features. We use Outer Product-based Neural Network (OPNN) \cite{pnn}, and follow the standard settings of \cite{pnn}, i.e., one embedding layer with the embedding dimension of 10, one product layer and three hidden layers of size 512, 256, 128 respectively where we set dropout rate at 0.5.
	\item The third dataset is the Criteo Display Ads dataset \cite{criteo} which contains approximately 46M samples over 7 days. There are 13 integer features and 26 categorical features. After one-hot encoding of categorical features, we have a total of 33.8M features. We split the dataset into 7 partitions in chronological order and select the earliest 6 parts for training which contains 29.6M features and the rest for testing though the dataset has no timestamp. We use Deep \& Cross Network (DCN) \cite{dcn} and choose the following settings\footnote{Limited by training resources available, we don't use the optimal hyperparameter settings of \cite{dcn}.}: one embedding layer with embedding dimension 8, two deep layers of size 64 each, and two cross layers.
\end{enumerate}
For the convenience of discussion, we use MLP, OPNN and DCN to represent the aforementioned three datasets coupled with their corresponding models. It is obvious that the embedding layer has most of parameters of the neural networks when the features have very high dimension, therefore we just add the regularization terms to the embedding layer. Furthermore, each embedding vector is considered as a group,  and a visual comparison between $\ell_{1}$, $\ell_{21}$ and mixed regularization effect is given in Fig. 2 of \cite{groupsparse}.

We treat the training set as the streaming data, hence we train 1 epoch with a batch size of 512 and do the validation. The experiments are conducted with 4-9 workers and 2-3 parameter servers in the TensorFlow framework \cite{tensorflow}, which depends on the different sizes of the datasets. According to \cite{auc}, area under the receiver-operator curve (AUC) is a good measurement in CTR estimation and AUC is widely adopted as the evaluation criterion in classification problems. Thus we choose AUC as our evaluation criterion.
We explore 5 learning rates from $1\text{e-5}$ to $1\text{e-1}$ with increments of 10$\times$ and choose the one with the best AUC for each new optimizer in the case of no regularization terms (It is equivalent to the original optimizer according to Theorem~\ref{thm:equivalent}). The details are listed in Table~\ref{tab:learning_rate} of Appendix~\ref{appendix:emp_rst}. All the experiments are run 5 times repeatedly and tested statistical significance using t-test. Without loss of generality, we choose two new optimizers to validate the performance, which are \textsc{Group Adam} and \textsc{Group AdaGrad}.

\subsection{\adam\ vs. \textsc{Group Adam}}
\label{subsec:4.2}

First, we compare the performance of the two optimizers on the same sparsity level. We set $\lambda_{1} = \lambda_{2} = 0$ and choose different values of $\lambda_{21}$ of Algorithm~\ref{alg:group-adam}, i.e., \textsc{Group Adam}, and achieve the same sparsity with \adam{} that uses the magnitude pruning method. Since we should delete the entire embedding vector which the feature corresponds to, not a single weight, and the amount of the features will dynamically increase as the training goes on, our method is different from the commonly used method \cite{prune}. Concretely, our method works in three steps. The first step sorts the norm of embedding vector from largest to smallest, and keeps top N embedding vectors which depend on the sparsity when finishing the first phase of training. In the second step we fine-tune the model. Since some new or deleted features will appear in the model after training with new data, in the last step we need to prune the model again to ensure that the desired sparsity is reached. We use the schedule of keeping 0\%, 10\%, 20\%, 30\% training samples to fine tune, and choose the best one. The details are listed in Table~\ref{tab:fine_tune} of Appendix~\ref{appendix:emp_rst}.

\begin{figure}[!tb]
	\centering
	\includegraphics[width=.32\linewidth]{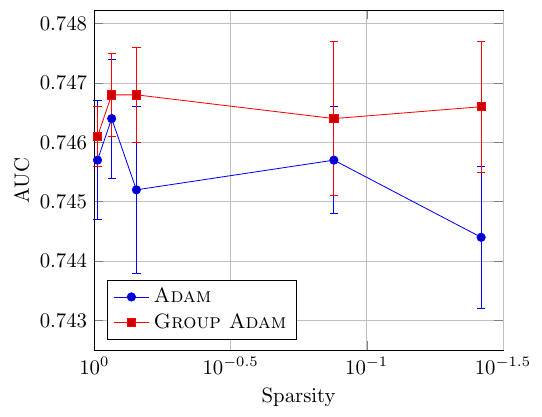}
	\includegraphics[width=.32\linewidth]{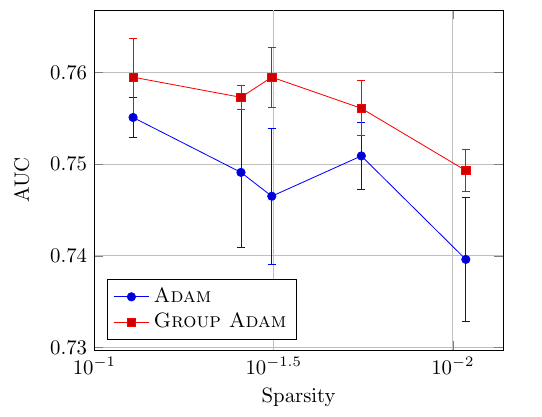}
	\includegraphics[width=.32\linewidth]{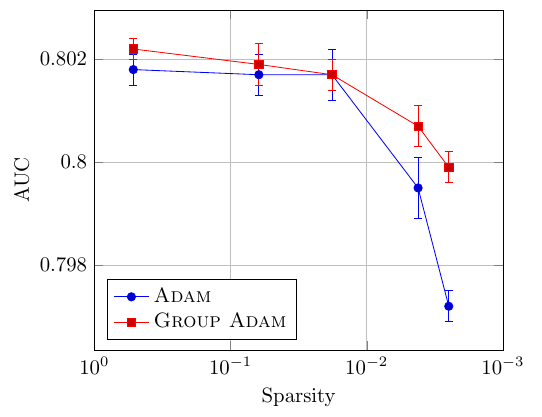}
	\caption{AUC across different sparsity on two optimizers for the three datasets. MLP, OPNN and DCN are in left, middle, right column respectively. The x-axis is sparsity (number of non-zero features whose embedding vectors are not equal to \bm{$0$} divided by the total number of features present in the training data). The y-axis is AUC. Error bars represent one standard deviation.}
	\label{fig:public_reg_loss}
	\vskip -0.15in
\end{figure}

Table~\ref{tab:reg_loss} reports the average results of the two optimizers in the three datasets. Note that \textsc{Group Adam} significantly outperforms \adam{} on the AUC metric in the same sparsity level for most experiments especially under extreme sparsity. (60\% experiments show statistically significant with 90\% confidence level, and 87.5\% experiments show statistically significant with 90\% confidence level when sparsity level is less than 5\%). Furthermore, as shown in Figure~\ref{fig:public_reg_loss}, the same $\ell_{21}$-regularization strength $\lambda_{21}$ has different effects of sparsity and accuracy on different datasets. The best choice of $\lambda_{21}$ depends on the dataset as well as the application (For example, if the memory of serving resource is limited, sparsity might be relatively more important). One can trade off accuracy to get more sparsity by increasing the value of $\lambda_{21}$.

\begin{table}[!tb]
	\caption{
		AUC for the two optimizers and sparsity (feature rate) in parentheses. The best AUC for each dataset on each sparsity level is bolded. The p-value of the t-test of AUC is also listed.
	}
	\small
	\label{tab:reg_loss}
	\centering
	\begin{threeparttable}
		\centering
		\begin{tabular}{cccccccccc}
			\toprule
			\bm{$\lambda_{21}$} & \multicolumn{3}{c}{\textbf{MLP}} & \multicolumn{3}{c}{\textbf{OPNN}} & \multicolumn{3}{c}{\textbf{DCN}}     \\ 
			\tiny \textsc{Group Adam} & \tiny \textsc{Adam} & \tiny \textsc{Group Adam} & \tiny P-Value & \tiny \textsc{Adam} & \tiny \textsc{Group Adam} & \tiny P-Value & \tiny \textsc{Adam} & \tiny \textsc{Group Adam} & \tiny P-Value \\
			\midrule
			$1\text{e-}4$ & \tabincell{c}{0.7457\\(0.974)} & \tabincell{c}{\textbf{0.7461} \\ (0.974)} & 0.470 & \tabincell{c}{0.7551\\(0.078)} & \tabincell{c}{\textbf{0.7595}\\(0.078)} & 0.086 & \tabincell{c}{0.8018\\(0.518)} & \tabincell{c}{\textbf{0.8022}\\(0.518)} & 0.105  \\ 
			\midrule
			$5\text{e-}4$ & \tabincell{c}{0.7464\\ (0.864) }& \tabincell{c}{\textbf{0.7468}\\ (0.864)} & 0.466 & \tabincell{c}{0.7491\\ (0.039) }& \tabincell{c}{\textbf{0.7573}\\ (0.039)} & 0.091 & \tabincell{c}{0.8017\\(0.062)} & \tabincell{c}{\textbf{0.8019}\\(0.062)} & 0.487 \\ 
			\midrule
			$1\text{e-}3$ & \tabincell{c}{0.7452\\ (0.701)} & \tabincell{c}{\textbf{0.7468}\\ (0.701) } & 0.058 & \tabincell{c}{0.7465\\ (0.032) }& \tabincell{c}{\textbf{0.7595}\\ (0.032)} & 0.014 & \tabincell{c}{0.8017\\(0.018)} & \tabincell{c}{0.8017\\(0.018)} & 0.943 \\ 
			\midrule
			$5\text{e-}3$ & \tabincell{c}{0.7457\\ (0.132)} & \tabincell{c}{\textbf{0.7464}\\ (0.132) } & 0.335 & \tabincell{c}{0.7509\\ (0.018) }& \tabincell{c}{\textbf{0.7561}\\ (0.018)} & 0.041 & \tabincell{c}{0.7995\\(4.2e-3)} & \tabincell{c}{\textbf{0.8007}\\(4.2e-3)} & 9.11e-3 \\ 
			\midrule
			$1\text{e-}2$ & \tabincell{c}{0.7444\\ (0.038)} & \tabincell{c}{\textbf{0.7466}\\ (0.038)} & 0.014 & \tabincell{c}{0.7396\\ (9.2e-3) }& \tabincell{c}{\textbf{0.7493}\\ (9.2e-3)} & 0.031 & \tabincell{c}{0.7972\\(2.5e-3)} & \tabincell{c}{\textbf{0.7999}\\(2.5e-3)} & 5.97e-7\\ 
			\bottomrule
		\end{tabular}
	\end{threeparttable}
	\vskip -0.1in
\end{table}

Next, we compare the performance of \adam{} without post-processing procedure, i.e., no magnitude pruning, and \textsc{Group Adam} under extremely high sparsity. We search regularization terms according to AUC and the values are listed in Table~\ref{tab:group_adam_hyp} of Appendix~\ref{appendix:emp_rst}. In general, good default settings of $\lambda_{2}$ is $1\text{e-5}$. The results are shown in Table~\ref{tab:best_auc}. Note that compared with \adam{}, \textsc{Group Adam} with appropriate regularization terms can achieve significantly better or highly competitive performance with producing extremely high sparsity.

\begin{table}[!tb]
	\small
	\caption{
		AUC for three datasets and sparsity (feature rate) in parentheses. The best value for each dataset is bolded. The p-value of t-test is also listed.
	}
	\label{tab:best_auc}
	\centering
	\begin{threeparttable}
		\centering
		\begin{tabular}{ccccccc}
			\toprule
			\textbf{Dataset} & \textbf{\adam} & \textbf{\textsc{Group Adam}} & \textbf{P-Value} & \textbf{\adagrad} & \textbf{\textsc{Group Adagrad}} & \textbf{P-Value} \\
			\midrule
			MLP & \tabincell{c}{0.7458\\ (1.000)} & \tabincell{c}{\textbf{0.7486}\\ (\textbf{0.018})} & \tabincell{c}{1.10e-3\\ (2.69e-11)} & \tabincell{c}{0.7453\\ (1.000)} & \tabincell{c}{\textbf{0.7469}\\ (\textbf{0.063})} & \tabincell{c}{0.106\\ (1.51e-9)} \\
			\midrule
			OPNN & \tabincell{c}{0.7588\\ (0.827)} & \tabincell{c}{\textbf{0.7617}\\ (\textbf{0.130})} & \tabincell{c}{0.289\\ (6.20e-11)} & \tabincell{c}{0.7556\\ (0.827)} & \tabincell{c}{\textbf{0.7595}\\ (\textbf{0.016})} & \tabincell{c}{0.026\\ ($<$ 2.2e-16)} \\
			\midrule
			DCN & \tabincell{c}{\textbf{0.8021}\\ (1.000)} & \tabincell{c}{0.8019\\ (\textbf{0.030})} & \tabincell{c}{0.422\\ (1.44e-11)} & \tabincell{c}{0.7975\\ (1.000)} & \tabincell{c}{\textbf{0.7978}\\ (\textbf{0.040})} & \tabincell{c}{0.198\\ (3.94e-11)} \\
			\bottomrule
		\end{tabular}
	\end{threeparttable}
	\vskip -0.15in
\end{table}

\subsection{\adagrad\  vs. \textsc{Group Adagrad}}
\label{subsec:4.3}

We compare the performance of \adagrad{} without magnitude pruning and \textsc{Group Adagrad} under extremely high sparsity. The regularization terms we choose are listed
in Table~\ref{tab:group_adagrad_hyp} of Appendix~\ref{appendix:emp_rst}. The results are also shown in Table~\ref{tab:best_auc}. Again note that in comparison to \adagrad{}, \textsc{Group Adagrad} can not only achieve significantly better or highly competitive performance of AUC, but also effectively and efficiently reduce the dimensions of the features.

\subsection{Discussion}
\label{subsec:4.4}

In this section we will compare the performance of $s_t$ with $\tilde{s}_t$ discussed in Section~\ref{subsec:2.1}, i.e., using $\tilde{s}_t$ means that replacing $\|s_t\|$ with $\|\tilde{s}_t\|$ in line 10 of Algorithm~\ref{alg:ada-group-lasso}. Furthermore, we will discuss the hyperparameters of $\ell_1$-regularization, $\ell_{21}$-regularization and emdedding dimension to show how these hyperparameters affect the effects of regularization. Without loss of generality, all experiments are conducted on DCN using \textsc{Group Adam}. The default settings of regularization terms are all zeros, unless otherwise specified.

\paragraph{\bm{$s_t$} vs. \bm{$\tilde{s}_t$}}
We choose $\ell_{21}$-regularization of $s_t$ and $\tilde{s}_t$ from 10 points in different sparsity levels. 
The details are listed in Table~\ref{tab:st_vs_stilde} of Appendix~\ref{appendix:emp_rst}. As shown in Figure~\ref{fig:st_vs_stilde}, the algorithm using $s_t$ outperforms the one using $\tilde{s}_t$ in the same level of sparsity.


\paragraph{\bm{$\ell_{1}$} vs. \bm{$\ell_{21}$}}
From lines 8 and 10 of Algorithm~\ref{alg:ada-group-lasso}, we know that if $z_t$ has the same elements, the values of $\ell_{1}$ and $\ell_{21}$, i.e., $\lambda_{1}$ and $\lambda_{21}$, have the same regularization effects. However, this situation almost cannot happen in reality. We compare the regularization performance with the same values of $\lambda_{1}$ and $\lambda_{21}$. The results are shown in Figure~\ref{fig:l1_vs_l21}. It is obvious that $\ell_{21}$-regularization is much more effective than $\ell_{1}$-regularization in producing sparsity. 
Therefore, if we just need to produce a sparse model, tuning $\lambda_{21}$ while keeping $\lambda_{1} = 0$ is usually a simple but effective choice.


\begin{figure}[!tb]
	\begin{minipage}[t]{0.48\columnwidth}
		\centering
		\includegraphics[width=\columnwidth]{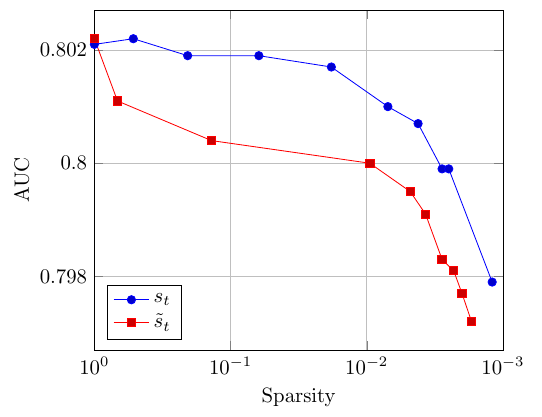}
		\caption{AUC across different sparsity (feature rate) on two methods. The legend is the algorithms using $s_t$ and $\tilde{s}_t$. The x-axis is sparsity. The y-axis is AUC.}
		\label{fig:st_vs_stilde}
	\end{minipage}
    \hskip 0.1in
	\begin{minipage}[t]{0.48\columnwidth}
		\centering
		\includegraphics[width=\columnwidth]{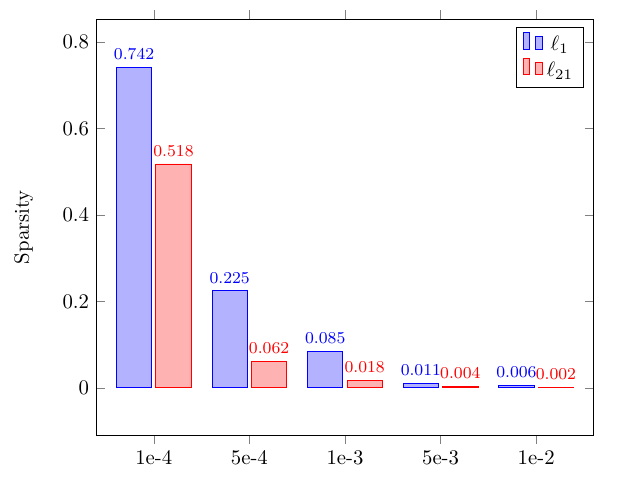}
		\caption{The sparsity (feature rate) across different values of regularized terms. The legend is the regularized terms. The x-axis is the values of regularized terms. The y-axis is sparsity.}
		\label{fig:l1_vs_l21}
	\end{minipage}
\end{figure}


\paragraph{Embedding Dimension}
Table~\ref{tab:emb_dim} reports the average results of different embedding dimensions, whose regularization terms are same to DCN of Table~\ref{tab:group_adam_hyp} of Appendix~\ref{appendix:emp_rst}. Note that the sparsity increases with the growth of the embedding dimension. The reason is that the square root of the embedding dimension is the multiplier of $\ell_{21}$-regularization.

\begin{table}[!tb]
	\small
	\caption{
		The sparsity (feature rate) for different embedding dimensions and AUC in parentheses. The best results are bolded.
	}
	\label{tab:emb_dim}
	\centering
	\begin{threeparttable}
		\centering
		\begin{tabular}{cc}
			\toprule
			\textbf{Embedding Dimension} & \textbf{\textsc{Group Adam}} \\
			\midrule
			4 & 0.074 (0.8008) \\
			8 & 0.030 (0.8019) \\
			16 & 0.012 (\textbf{0.8020}) \\
			32 & \textbf{0.008} (0.8011) \\
			\bottomrule
		\end{tabular}
	\end{threeparttable}
	\vskip -0.15in
\end{table}

\section{Conclusion}
\label{sec:5}

In this paper, we propose a novel framework that adds the regularization terms to a family of adaptive optimizers for producing sparsity of DNN models. We apply this framework to create a new class of optimizers. We provide closed-form solutions and algorithms with slight modification. We built the relation between new and original optimizers, i.e., our new optimizers become equivalent with the corresponding original ones, once the regularization terms vanish. We theoretically prove the convergence rate of the regret and also conduct empirical evaluation on the proposed optimizers in comparison to the original optimizers with and without magnitude pruning. The results clearly demonstrate the advantages of our proposed optimizers in both getting significantly better performance and producing sparsity. Finally, it would be interesting in the future to investigate the convergence in non-convex settings and evaluate our optimizers on more applications from fields such as compute vision, natural language processing and etc.

\appendix
\clearpage
\section*{Appendix}

\section{Proof of Theorem~\ref{thm:ada-group-lasso}}
\label{appendix:ada-group-lasso}

\begin{proof}
	\begin{equation}
	\begin{aligned}
	x_{t+1} =& \argmin_{x} m_{1:t} \cdot x + \sum_{s=1}^{t}\frac{1}{2\alpha_{s}}(x - x_s)^T Q_s (x - x_s) \\
	& + \Psi_t(x) \\
	=& \argmin_{x} m_{1:t} \cdot x + \sum_{s=1}^{t}\frac{1}{2\alpha_s}\big( \|Q_s^{\frac{1}{2}}x\|_2^2-2x\cdot (Q_s x_s)\ \\
	& + \|Q_s^{\frac{1}{2}}x_s\|_2^2\big) + \Psi_t(x) \\
	=& \argmin_{x}{ } ( m_{1:t} - \sum_{s=1}^{t}\frac{Q_s}{\alpha_s}x_s) \cdot x + \sum_{s=1}^{t}\frac{1}{2\alpha_s}\|Q_s^{\frac{1}{2}}x\|_2^2 \\
	& + \Psi_t(x).
	\end{aligned}
	\label{eq:xt}
	\end{equation}
	Define $z_{t-1} = m_{1:t-1} - \sum_{s=1}^{t-1}\frac{Q_s}{\alpha_s}x_s$ ($t\geq 2$) and we can calculate $z_t$ as
	\begin{align}
	z_t = z_{t-1} + m_t - \frac{Q_t}{\alpha_t}x_t, \quad t\geq 1.
	\label{eq:zt}
	\end{align}
	By substituting \eqref{eq:zt}, \eqref{eq:xt} is simplified to be
	\begin{align}
	x_{t+1} = \argmin_{x} z_t \cdot x + \sum_{s=1}^{t}\frac{Q_s}{2\alpha_s}\|x\|_2^2 + \Psi_t(x).
	\label{eq:xt_2}
	\end{align}
	By substituting $\Psi_t(x)$ (Eq.~\eqref{eq:sgl}) into \eqref{eq:xt_2}, we get
	\begin{equation}
	\begin{aligned}
	x_{t+1} =& \argmin_{x} z_t \cdot x + \sum_{g=1}^{G} \big( \lambda_{1}\|x^g\|_{1} + \lambda_{21}\sqrt{d_{x^g}} \\ &\|(\sum_{s=1}^{t}\frac{Q_s^g}{2\alpha_{s}} + \lambda_{2}\mathbb{I})^{\frac{1}{2}}x^g\|_{2} \big) +  \|(\sum_{s=1}^{t}\frac{Q_s}{2\alpha_s} + \lambda_{2}\mathbb{I})^{\frac{1}{2}}x\|_2^2.
	\end{aligned}
	\label{eq:xt_3}
	\end{equation}
	\noindent Since the objective of \eqref{eq:xt_3} is component-wise and element-wise, we can focus on the solution in one group, say $g$, and one entry, say $i$, in the $g$-th group. Let $\sum_{s=1}^{t}\frac{Q_s^g}{2\alpha_{s}} = \diag(\sigma_t^g)$ where $\sigma_t^g = (\sigma^g_{t,1}, \dots, \sigma^g_{t, d_{x^g}})$. The objective of \eqref{eq:xt_3} on $x_{t+1,i}^{g}$ is
	\begin{equation}
	\Omega(x_{t+1,i}^{g}) = z_{t,i}^{g} x_{t+1,i}^{g} + \lambda_{1}|x_{t+1,i}^{g}| + \Phi(x_{t+1,i}^{g}),
	\label{eq:xt_4}
	\end{equation}
	where $\Phi(x_{t+1,i}^{g}) = \lambda_{21}\sqrt{d_{x^g}}\|(\sigma^{g}_{t,i} + \lambda_{2})^\frac{1}{2} x^{g}_{t+1,i}\|_2 + \|(\sigma^{g}_{t,i} + \lambda_{2})^\frac{1}{2} x^{g}_{t+1,i}\|_2^2$ is a non-negative function and $\Phi(x_{t+1,i}^{g}) = 0$ iff $x_{t+1,i}^{g} = 0$ for all $i \in \{1, \dots, d_{x^g}\}$.
	
	\noindent We discuss the optimal solution of \eqref{eq:xt_4} in three cases:
	\begin{enumerate}[a)]
		\item If $z_{t,i}^{g} = 0$, then $x_{t+1,i}^{g} = 0$.
		\item If $z_{t,i}^{g} > 0$, then $x_{t+1,i}^{g} \leq 0$. Otherwise, if $x_{t+1,i}^{g} > 0$, we have $\Omega(-x_{t+1,i}^{g}) < \Omega(x_{t+1,i}^{g})$, which contradicts the minimization value of $\Omega(x)$ on $x_{t+1,i}^{g}$. 
		
		Next, if $z_{t,i}^{g} \leq \lambda_{1}$, then $x_{t+1,i}^{g} = 0$. Otherwise, if $x_{t+1,i}^{g} < 0$, we have $\Omega(x_{t+1,i}^{g}) = (z_{t,i}^{g} -  \lambda_{1})x_{t+1,i}^{g} + \Phi(x_{t+1}^{g,i}) > \Omega(0)$, which also contradicts the minimization value of $\Omega(x)$ on $x_{t+1,i}^{g}$.
		
		Third, $z_{t,i}^{g} > \lambda_{1}\ (\forall\ i = 1, \dots, d_{x^g})$. The objective of \eqref{eq:xt_4} for the $g$-th group, $\Omega(x_{t+1}^g)$, becomes 
		$$(z_t^g - \lambda_{1} \boldsymbol{1}_{d_{x^g}})\cdot x_{t+1}^g + \Phi(x_{t+1}^g).$$
		\item If $z_{t,i}^{g} < 0$, the analysis is similar to b). We have $x_{t+1,i}^{g} \geq 0$. When $-z_{t,i}^{g} \leq \lambda_{1}$, $x_{t+1,i}^{g} = 0$. When $-z_{t,i}^{g} > \lambda_{1}\ (\forall\ i = 1, \dots, d_{x^g})$, we have
		$$\Omega(x_{t+1}^g) = (z_t^g + \lambda_{1} \boldsymbol{1}_{d_{x^g}})\cdot x_{t+1}^g + \Phi(x_{t+1}^g).$$
	\end{enumerate}
	
	\noindent From a), b), c) above, we have
	\begin{equation}
	x_{t+1}^g = \argmin_{x} -s_t^g \cdot x + \Phi(x),
	\label{eq:xt_5}
	\end{equation}
	where the $i$-th element of $s_t^g$ is defined same as \eqref{eq:s_t}.
	
	\noindent Define 
	\begin{equation}
	y = (\diag(\sigma^{g}_t) + \lambda_{2}\mathbb{I})^\frac{1}{2} x.
	\label{eq:yt}
	\end{equation}
	
	\noindent By substituting \eqref{eq:yt} into \eqref{eq:xt_5}, we get
	\begin{equation}
	y_{t+1}^g = \argmin_{y} -\tilde{s}_t^g \cdot y + \lambda_{21}\sqrt{d_{x^g}}\|y\|_2 + \|y\|_2^2,
	\label{eq:yt_2}
	\end{equation}
	where $\tilde{s}_t^g = (\diag(\sigma^g_t) + \lambda_{2}\mathbb{I})^{-1} s_t^g$ which is defined same as \eqref{eq:st_2}. This is unconstrained non-smooth optimization problem. Its optimality condition (see \cite{convex}, Section 27) states that $y_{t+1}^g$ is an optimal solution if and only if there exists $\xi \in \partial\|y_{t+1}^g\|_2$ such that 
	\begin{equation}
	-\tilde{s}_t^g + \lambda_{21}\sqrt{d_{x^g}} \xi + 2y_{t+1}^g = 0.
	\label{eq:yt_3}
	\end{equation}
	The subdifferential of $\|y\|_2$ is 
	\begin{equation*}
	\partial\|y\|_2 = \left\{ 
	\begin{array}{ll}
	\{ \zeta \in \mathbb{R}^{d_{x^g}} | -1 \leq \zeta^{(i)} \leq 1, i = 1, \dots, d_{x^g} \} & \textrm{if $y = 0$,}\\
	\frac{y}{\|y\|_2} & \textrm{if $y \neq 0$.}
	\end{array} \right.
	\label{eq:yt_4}
	\end{equation*}
	Similarly to the analysis of $\ell_{1}$-regularization, we discuss the solution of \eqref{eq:yt_3} in two different cases:
	\begin{enumerate}[a)]
		\item If $\|\tilde{s}_t^g\|_2 \leq \lambda_{21}\sqrt{d_{x^g}}$, then $y_{t+1}^g = 0$ and $\xi = \frac{\tilde{s}_t^g}{\lambda_{21}\sqrt{d_{x^g}}} \in \partial\|0\|_2$ satisfy \eqref{eq:yt_3}. We also show that there is no solution other than $y_{t+1}^g = 0$. Without loss of generality, we assume $y_{t+1,i}^{g} \neq 0$ for all $i \in \{1, \dots, d_{x^g}\}$, then $\xi = \frac{y_{t+1}^{g}}{\|y_{t+1}^g\|_2}$, and
		\begin{equation}
		-\tilde{s}_t^g + \frac{\lambda_{21}\sqrt{d_{x^g}}}{\|y_{t+1}^g\|_2} y_{t+1}^g + 2y_{t+1}^g = 0.
		\label{eq:yt_5}
		\end{equation}
		From \eqref{eq:yt_5}, we can derive
		\begin{equation*}
		(\frac{\lambda_{21}\sqrt{d_{x^g}}}{\|y_{t+1}^g\|_2} + 2) \|y_{t+1}^g\|_2 = \|\tilde{s}_t^g\|_2.
		\end{equation*}
		Furthermore, we have 
		\begin{equation}
		\|y_{t+1}^g\|_2 = \frac{1}{2}(\|\tilde{s}_t^g\|_2 - \lambda_{21}\sqrt{d_{x^g}}), 
		\label{eq:yt_6}
		\end{equation}
		where $\|y_{t+1}^g\|_2 > 0$ and $\|\tilde{s}_t^g\|_2 - \lambda_{21}\sqrt{d_{x^g}} \leq 0$ contradict each other.
		\item If $\|\tilde{s}_t^g\|_2 > \lambda_{21}\sqrt{d_{x^g}}$, then from \eqref{eq:yt_5} and \eqref{eq:yt_6}, we get 
		\begin{equation}
		y_{t+1}^g = \frac{1}{2}(1 - \frac{\lambda_{21}\sqrt{d_{x^g}}}{\|\tilde{s}_t^g\|_2})\tilde{s}_t^g.
		\label{eq:yt_7}
		\end{equation}
		We replace $y_{t+1}^g$ of \eqref{eq:yt_7} by $x_{t+1}^g$ using \eqref{eq:yt}, then we have
		\begin{equation}
		\begin{aligned}
		x_{t+1}^g &= (\diag(\sigma_t^g) + \lambda_{2}\mathbb{I})^{-\frac{1}{2}} y_{t+1}^g \\
		&= (2\diag(\sigma_t^g) + 2\lambda_{2}\mathbb{I})^{-1} (1 - \frac{\lambda_{21}\sqrt{d_{x^g}}}{\| \tilde{s}_t^g\|_2}) s_t^g \\
		&= (\sum_{s=1}^{t} \frac{Q_s}{\alpha_{s}} + 2\lambda_{2}\mathbb{I})^{-1} (1 - \frac{\lambda_{21}\sqrt{d_{x^g}}}{\| \tilde{s}_t^g\|_2}) s_t^g.	   
		\end{aligned}
		\end{equation}
	\end{enumerate}
	Combine a) and b) above, we finish the proof.
\end{proof}

\section{Proof of Theorem~\ref{thm:equivalent}}
\label{appendix:equivalent}

	\begin{proof}
	We use the method of induction.
	\begin{enumerate}[a)]
		\item When $t = 1$, then Algorithm~\ref{alg:ada-group-lasso} becomes 
		\begin{equation*}
		\begin{aligned}
		Q_1 &= \alpha_1(\frac{\sqrt{V_1}}{\alpha_1} - \frac{\sqrt{V_0}}{\alpha_0}) = \sqrt{V_1}, \\
		z_1 &= z_0 + m_1 - \frac{Q_1}{\alpha_1} x_1 = m_1 -  \frac{\sqrt{V_1}}{\alpha_1} x_1, \\
		s_1 &= -z_1 = \frac{\sqrt{V_1}}{\alpha_1} x_1 - m_1, \\
		x_2 &= (\frac{\sqrt{V_1}}{\alpha_1})^{-1}s_1 = x_1 - \alpha_1 \frac{m_1}{\sqrt{V_1}},
		\end{aligned}
		\end{equation*}
		which equals to Eq. \eqref{eq:general-formula}.
		\item Assume $t = T$, Eq.~\eqref{eq:equal}  are true. 
		\begin{equation}
		\begin{aligned}
		z_T &= m_T - \frac{\sqrt{V_T}}{\alpha_{T}}x_T, \\
		x_{T+1} &= x_T - \alpha_{T}\frac{m_T}{\sqrt{V_T}}.
		\label{eq:equal}
		\end{aligned}
		\end{equation}
		For $t = T+1$, we have 
		\begin{equation*}
		\begin{aligned}
		z_{T+1} &= z_T + m_{T+1} - \frac{Q_{T+1}}{\alpha_{T+1}} x_{T+1} \\
		&= m_T - \frac{\sqrt{V_T}}{\alpha_{T}}x_T + m_{T+1} - \frac{Q_{T+1}}{\alpha_{T+1}} x_{T+1} \\
		&=m_T - \frac{\sqrt{V_T}}{\alpha_{T}}(x_{T+1} + \alpha_{T} \frac{m_T}{\sqrt{V_T}}) + m_{T+1} - \frac{Q_{T+1}}{\alpha_{T+1}} x_{T+1} \\
		&= m_{T+1} - (\frac{\sqrt{V_T}}{\alpha_{T}} + \frac{Q_{T+1}}{\alpha_{T+1}}) x_{T+1} \\
		&= m_{T+1} - \frac{\sqrt{V_{T+1}}}{\alpha_{T+1}} x_{T+1}, \\
		x_{T+2} &= (\frac{\sqrt{V_{T+1}}}{\alpha_{T+1}})^{-1} s_{T+1}  = -(\frac{\sqrt{V_{T+1}}}{\alpha_{T+1}})^{-1} z_{T+1} \\
		&= x_{T+1} - \alpha_{T} \frac{m_{T+1}}{\sqrt{V_{T+1}}}.
		\end{aligned}
		\end{equation*}	
	\end{enumerate}
	Hence, we complete the proof.
\end{proof}

\section{Proof of Theorem~\ref{thm:convergence}}
\label{appendix:convergence}

	\begin{proof}
	Let 
	\begin{equation*}
	h_t(x) = \left\{ 
	\begin{array}{ll}
	\sum_{s=1}^{t} \frac{1}{2\alpha_{s}} \| Q_s^{\frac{1}{2}}(x - x_s)\|_2^2 & \forall\ t\in[T],\\
	\frac{1}{2}\|x-c\|_2^2 & t=0.
	\end{array}
	\right.
	\end{equation*}
	It is easy to verify that for all $t\in{}[T]$, $h_t(x)$ is $1$-strongly convex with respect to $\|\cdot\|_{\sqrt{V_t}/\alpha_{t}}$ which $\frac{\sqrt{V_t}}{\alpha_{t}} = \sum_{s=1}^{t} \frac{Q_s}{\alpha_{s}}$ , and $h_0(x)$ is $1$-strongly convex with respect to $\|\cdot\|_2$.
	
	\noindent From \eqref{eq:regret}, we have
	
	
	\begin{equation}
	\begin{aligned}
	\mathcal{R}_{T} =& \sum_{t=1}^{T} (f_t(x_t) - f_t(x^*)) \leq \sum_{t=1}^{T} \left<g_t, x_t - x^* \right> \\
	=& \sum_{t=1}^{T} \left<m_t - \gamma m_{t - 1}, x_t - x^* \right> \leq \sum_{t=1}^{T} \left<m_t, x_t - x^* \right> \\
	=& \sum_{t=1}^{T} \left<m_t, x_t \right> + \Psi_T(x^*) + h_T(x^*) + \big(\sum_{t=1}^{T} \left<-m_t, x^*\right> \\
	&- \Psi_T(x^*) - h_T(x^*)\big) \\
	\leq& \sum_{t=1}^{T}\left<m_t, x_t \right> + \Psi_T(x^*) + h_T(x^*) + \sup_{x\in\mathcal{Q}} \big\{ \left<-m_{1:T}, x \right> \\
	&- \Psi_T(x) - h_T(x) \big\},
	\end{aligned}
	\label{eq:rt}
	\end{equation}
	where in the first and second inequality above, we use the convexity of $f_t(x)$ and the condition \eqref{eq:cond} respectively.
	
	\noindent We define $h_t^*(u)$ to be the conjugate dual of $\Psi_t(x) + h_t(x)$:
	$$h_t^*(u) = \sup_{x\in\mathcal{Q}}\left\{\left<u, x\right> - \Psi_t(x) - h_t(x)\right\},\quad t\geq 0,$$
	where $\Psi_0(x) = 0$.
	Since $h_t(x)$ is $1$-strongly convex with respect to the norm $\|\cdot\|_{h_t}$, the function $h_t^*$ has $1$-Lipschitz continuous gradients with respect to $\|\cdot\|_{h_t^*}$ (see, \cite{primaldual2}, Theorem 1):
	\begin{equation}
	\|\nabla h_t^*(u_1) - \nabla h_t^*(u_2)\|_{h_t} \leq \|u_1 - u_2\|_{h_t^*},
	\label{eq:ht}
	\end{equation}
	and 
	\begin{equation}
	\nabla h_t^*(u) = \argmin_{x\in\mathcal{Q}} \left\{ -\left<u, x\right> + \Psi_t(x) + h_t(x)\right\}.
	\label{eq:ht_2}
	\end{equation}
	As a trivial corollary of \eqref{eq:ht}, we have the following inequality:
	\begin{equation}
	h_t^*(u + \delta) \leq h_t^*(u) + \left<\nabla h_t^*(u), \delta\right> + \frac{1}{2} \|\delta\|_{h_t^*}^2.
	\label{eq:ht_3}
	\end{equation}
	Since $h_{t+1}(x) \geq h_t(x)$ and $\Psi_{t+1}(x) \geq \Psi_t(x)$, from \eqref{eq:ht_2}, \eqref{eq:ht_3}, \eqref{eq:ada-sgl}, we have
	\begin{equation}
	\begin{aligned}
	h_T^*(-m_{1:T}) \leq& h_{T-1}^*(-m_{1:T}) \\
	\leq& h_{T-1}^*(-m_{1:T-1}) - \left<\nabla h_{T-1}^*(-m_{1:T-1}), m_T\right>\\ 
	&+ \frac{1}{2}\|m_T\|_{h_{T-1}^*}^2 \\
	\leq& h_{T-2}^*(-m_{1:T-1}) - \left<x_T, m_T\right> + \frac{1}{2}\|m_T\|_{h^*_{T-1}}^2 \\
	\leq& h_0^*(0) - \left<\nabla h_0^*(0), m_1\right> - \sum_{t=2}^{T}\left<x_t, m_t\right> \\
	&+ \frac{1}{2}\sum_{t=2}^{T} \|m_t\|_{h_{t-1}^*}^2 \\
	=& -\sum_{t=1}^{T}\left<x_t, m_t\right> + \frac{1}{2}\sum_{t=1}^{T} \|m_t\|_{h_{t-1}^*}^2.
	\end{aligned}
	\label{eq:ht_4}
	\end{equation}
	where the last equality above follows from $h_0^*(0)=0$ and \eqref{eq:x1} which deduces $x_1 = \nabla h_0^*(0)$.
	
	\noindent By substituting \eqref{eq:ht_4}, \eqref{eq:rt} becomes
	\begin{equation}
	\begin{aligned}
	\mathcal{R}_T &\leq \sum_{t=1}^{T} \left<m_t, x_t \right> + \Psi_T(x^*) + h_T(x^*) + h_T^*(-m_{1:T}) \\
	&\leq \Psi_T(x^*) + h_T(x^*) + \frac{1}{2}\sum_{t=1}^{T} \|m_t\|_{h_{t-1}^*}^2.
	\end{aligned}
	\end{equation}
\end{proof}

\section{Additional Proofs}
\label{appendix:add_prof}
\subsection{Proof of Lemma~\ref{lem:1}}
\label{appendix:lemma1}

\begin{proof}
	Let $V_t = \diag(\sigma_t)$ where $\sigma_t$ is the vector of the diagonal elements of $V_t$. For $i$-th entry of $\sigma_t$, by substituting \eqref{eq:mt} into \eqref{eq:vt}, we have
	\begin{equation}
	\begin{aligned}
	\sigma_{t,i} &= g_{t,i}^2 + \eta\sigma_{t-1,i} \\
	&= (m_{t,i} - \gamma m_{t-1,i})^2 + \eta g_{t-1,i}^2 + \eta^2\sigma_{t-2, i} \\
	&=\sum_{s=1}^{t}\eta^{t-s}(m_{s,i} - \gamma m_{s-1,i})^2 \\
	&\geq \sum_{s=1}^{t}\eta^{t-s}(1-\gamma)(m_{s,i}^2 - \gamma m_{s-1,i}^2) \\
	&=(1 - \gamma) \big(m_{t,i}^2 + (\eta -\gamma)\sum_{s=1}^{t-1}\eta^{t-s-1} m_{s,i}^2\big).
	\end{aligned}
	\label{eq:vt_2}
	\end{equation}
	Next, we will discuss the value of $\eta$ in two cases.
	\begin{enumerate}[a)]
		\item $\eta = 1$.  From \eqref{eq:vt_2}, we have
		\begin{equation}
		\begin{aligned}
		\sigma_{t,i} &\geq (1 -\gamma) \big(m_{t,i}^2 + (1-\gamma)\sum_{s=1}^{t-1} m_{s,i}^2\big) \\
		&> (1-\gamma)^2 \sum_{s=1}^{t} m_{s,i}^2 \\
		&\geq (1-\nu)^2 \sum_{s=1}^{t} m_{s,i}^2.
		\label{eq:vt_3}
		\end{aligned}
		\end{equation}
		Recalling the definition of $M_{t,i}$ in Section~\ref{subsec:1.5}, from \eqref{eq:vt_3}, we have
		\begin{equation*}
		\sum_{t=1}^{T} \frac{m_{t,i}^2}{\sqrt{\sigma_{t,i}}} < \frac{1}{1-\nu}\sum_{t=1}^{T} \frac{m_{t,i}^2}{\|M_{t,i}\|_2} \leq \frac{2}{1-\nu} \|M_{T,i}\|_2,
		\end{equation*}
		where the last inequality above follows from Appendix C of \cite{adagrad}. Therefore, we get 
		\begin{equation}
		\begin{aligned}
		\sum_{t=1}^{T}\|m_t\|^2_{(\frac{\sqrt{V_t}}{\alpha_{t}})^{-1}} &= \alpha \sum_{t=1}^{T}\sum_{i=1}^{d} \frac{m_{t,i}^2}{\sqrt{\sigma_{t, i}}} \\
		&< \frac{2\alpha}{1-\nu} \sum_{i=1}^{d} \|M_{T, i}\|_2.
		\end{aligned}
		\label{eq:vt_4.1}
		\end{equation} 
		
		\item $\eta < 1$. We assume $\eta\geq\gamma$ and $\kappa V_t \succeq V_{t-1}$ where $\kappa < 1$, then we have 
		\begin{equation*}
		\sum_{s=1}^{t} \kappa^{t-s} \sigma_{t,i} \geq \sum_{s=1}^{t} \sigma_{s,i} \geq (1 - \gamma)\sum_{s=1}^{t} m_{s,i}^2.
		\end{equation*}
		Hence, we get 
		\begin{equation}
		\begin{aligned}
		\sigma_{t, i} &\geq \frac{1-\kappa}{1-\kappa^t}(1-\gamma) \sum_{s=1}^{t} m_{s,i}^2 \\
		&> (1 - \kappa)(1-\gamma) \sum_{s=1}^{t} m_{s,i}^2 \\
		&\geq (1 - \nu)^2 \sum_{s=1}^{t} m_{s,i}^2,
		\end{aligned}
		\end{equation}
		which deduces the same conclusion \eqref{eq:vt_4.1} of a).
	\end{enumerate}
	Combine a) and b), we complete the proof.
\end{proof}

\subsection{Proof of Corollary~\ref{coro:1}}
\label{appendix:coro1}

\begin{proof}
	From the definition of $m_t$ \eqref{eq:mt}, $V_t$ \eqref{eq:vt}, we have
	\begin{equation*}
	\begin{aligned}
	|m_{t, i}| &= |\sum_{s=1}^{t} \gamma^{t-s} g_{s,i}| \leq \frac{1-\gamma^t}{1-\gamma}G < \frac{G}{1-\gamma} \leq \frac{G}{1-\nu}, \\
	|\sigma_{t, i}| &= |\sum_{s=1}^{t} \eta^{t-s} g_{s,i}^2| \leq t G^2.
	\end{aligned}
	\end{equation*}
	Hence, we have 
	\begin{equation}
	\Psi_T(x^*) \leq \lambda_{1}dD_1 + \lambda_{21}dD_1(\frac{\sqrt{T}G}{2\alpha} + \lambda_{2})^{\frac{1}{2}} + \lambda_{2}dD_1^2,
	\label{eq:vt_8}	
	\end{equation}
	\begin{equation}
	h_T(x^*) \leq \frac{dD_2^2 G}{2\alpha} \sqrt{T},
	\label{eq:vt_9}
	\end{equation}
	\begin{equation}
	\frac{1}{2} \sum_{t=1}^{T}\|m_t\|_{h_{t-1}^*}^2 < \frac{\alpha}{1-\nu} \sum_{i=1}^{d} \frac{\sqrt{T}G}{1-\nu} = \frac{d\alpha G}{(1-\nu)^{2}}\sqrt{T}.
	\label{eq:vt_10}
	\end{equation}
	Combining \eqref{eq:vt_8}, \eqref{eq:vt_9}, \eqref{eq:vt_10}, we complete the proof.
\end{proof}

\section{Additional Experimental Results}
\label{appendix:emp_rst}

\begin{table}[!htbp]
	\small
	\caption{
		The learning rates of the optimizers of three datasets.
	}
	\label{tab:learning_rate}
	\centering
	\begin{threeparttable}
		\centering
		\begin{tabular}{cccc}
			\toprule
			\textbf{Optimizer} & \textbf{MLP} & \textbf{OPNN} & \textbf{DCN} \\
			\midrule
			\textsc{Adam/Group Adam} & $1\text{e-4}$ & $1\text{e-4}$ & $1\text{e-3}$ \\
			\textsc{AdaGrad/Group AdaGrad} & $1\text{e-2}$ & $1\text{e-2}$ & $1\text{e-2}$ \\
			\bottomrule
		\end{tabular}
	\end{threeparttable}
	\vskip -0.15in
\end{table}

\begin{figure}[!htbp]
	\begin{minipage}[t]{.48\columnwidth}
		\begin{table}[H]
			\small
			\caption{
				The regularization terms of \textsc{Group Adam} of three datasets.
			}
			\label{tab:group_adam_hyp}
			\centering
			\begin{threeparttable}
				\centering
				\begin{tabular}{cccc}
					\toprule
					\textbf{Dataset} & \bm{$\lambda_{1}$} & \bm{$\lambda_{21}$} & \bm{$\lambda_{2}$}\\
					\midrule
					MLP & $5\text{e-3}$ & $1\text{e-2}$ & $1\text{e-5}$ \\
					OPNN & $8\text{e-5}$ & $1\text{e-5}$ & $1\text{e-5}$ \\
					DCN & $4\text{e-4}$ & $5\text{e-4}$ & $1\text{e-5}$  \\
					\bottomrule
				\end{tabular}
			\end{threeparttable}
		\end{table}
	\end{minipage}
\hskip 0.1in
	\begin{minipage}[t]{.48\columnwidth}
		\begin{table}[H]
			\small
			\caption{
				The regularization terms of \textsc{Group Adagrad} of three datasets.
			}
			\label{tab:group_adagrad_hyp}
			\centering
			\begin{threeparttable}
				\centering
				\begin{tabular}{cccc}
					\toprule
					\textbf{Dataset} & \bm{$\lambda_{1}$} & \bm{$\lambda_{21}$} & \bm{$\lambda_{2}$}\\
					\midrule
					MLP & 0 & $1\text{e-2}$ & $1\text{e-5}$ \\
					OPNN & $8\text{e-5}$ & $8\text{e-5}$ & $1\text{e-5}$ \\
					DCN & 0 & $4\text{e-3}$ & $1\text{e-5}$  \\
					\bottomrule
				\end{tabular}
			\end{threeparttable}
		\end{table}
	\end{minipage}
\end{figure}

\begin{table}[!tb]
    \small
	\caption{
		Fine-tuned schedule of magnitude pruning. The best AUC for each dataset on each sparsity level is bolded.
	}
	\label{tab:fine_tune}
	\centering
	\renewcommand\tabcolsep{4.0pt}
	\begin{threeparttable}
		\centering
		\begin{tabular}{ccccccc}
			\toprule
			\multirow{2}{35pt}{\small \textbf{Ratio of Samples}} & \multicolumn{2}{c}{\textbf{MLP}} & \multicolumn{2}{c}{\textbf{OPNN}} & \multicolumn{2}{c}{\textbf{DCN}}     \\ 
			  & \tiny Sparsity & \tiny AUC & \tiny Sparsity & \tiny AUC & \tiny Sparsity & \tiny AUC \\
			\midrule
			30\% & \multirow{4}{*}{0.974} & 0.7454 & \multirow{4}{*}{0.078} & 0.6809 & \multirow{4}{*}{0.518} & 0.8016 \\
			20\% &  & \textbf{0.7457} &  & 0.7259 &  & 0.8015 \\
			10\% &  & 0.7439 &  & 0.7160 &  & 0.8015 \\
			0\% &  & 0.7452 &  & \textbf{0.7551} &  & \textbf{0.8019} \\
			
			\midrule
			30\% & \multirow{4}{*}{0.864} & 0.7454 & \multirow{4}{*}{0.039} & 0.6356 & \multirow{4}{*}{0.062} & 0.8005 \\
			20\% &  & 0.7441 &  & 0.6383 &  & 0.7977 \\
			10\% &  & 0.7453 &  & 0.6678 &  & 0.7884 \\
			0\% &  & \textbf{0.7464} &  & \textbf{0.7491} &  & \textbf{0.8018} \\
			
			\midrule
			30\% & \multirow{4}{*}{0.701} & 0.7443 & \multirow{4}{*}{0.032} & 0.6826 & \multirow{4}{*}{0.018} & 0.7883 \\
			20\% &  & 0.7452 &  & 0.6618 &  & 0.7969 \\
			10\% &  & 0.7449 &  & 0.6604 &  & 0.7800 \\
			0\% &  & \textbf{0.7452} &  & \textbf{0.7465} &  & \textbf{0.8017} \\
			
			\midrule
			30\% & \multirow{4}{*}{0.132} & \textbf{0.7457} & \multirow{4}{*}{0.018} & 0.6318 & \multirow{4}{*}{0.0042} & 0.7400 \\
			20\% &  & 0.7428 &  & 0.6419 &  & 0.7236 \\
			10\% &  & 0.7450 &  & 0.6207 &  & 0.6864 \\
			0\% &  & 0.7452 &  & \textbf{0.7509} &  & \textbf{0.7995} \\
			
			\midrule
			30\% & \multirow{4}{*}{0.038} & \textbf{0.7444} & \multirow{4}{*}{0.0092} & 0.5934 & \multirow{4}{*}{0.0025} & 0.7442 \\
			20\% &  & 0.7442 &  & 0.6003 &  & 0.7105 \\
			10\% &  & 0.7437 &  & 0.6355 &  & 0.6925 \\
			0\% &  & 0.7430 &  & \textbf{0.7396} &  & \textbf{0.7972} \\
			
			\bottomrule
		\end{tabular}
	\end{threeparttable}
\end{table}

\begin{table}[!tb]
	\small
	\caption{
		The $\ell_{21}$-regularization terms of the optimizer using $s_t$ and $\tilde{s}_t$.
	}
	\label{tab:st_vs_stilde}
	\centering
	\begin{threeparttable}
		\centering
		\begin{tabular}{cccccc}
			\toprule
			\textbf{Method} & \multicolumn{4}{c}{\bm{$\lambda_{21}$}} \\
			\midrule
			\multirow{2}{*}{$s_t$ } & 0 & 1e-4 & 2.5e-4 &  5e-4 &  1e-3 \\
											    & 2.5e-3 & 5e-3 & 7.5e-3 & 1e-2 & 2.5e-2  \\
			\midrule
			\multirow{2}{*}{$\tilde{s}_t$} & 0 & 0.05 & 0.075 & 0.1 & 0.125 \\
														 & 0.15 & 0.175 & 0.2 & 0.225 & 0.25 \\
			\bottomrule
		\end{tabular}
	\end{threeparttable}
	\vskip -0.1in
\end{table}


\begin{thebibliography}{8}
\providecommand{\url}[1]{\texttt{#1}}
\providecommand{\urlprefix}{URL }
\providecommand{\doi}[1]{https://doi.org/#1}

\bibitem{tensorflow}
Abadi, M., Barham, P., Chen, J., Chen, Z., Davis, A., Dean, J., Devin, M.,
Ghemawat, S., Irving, G., Isard, M., Kudlur, M., Levenberg, J., Monga, R.,
Moore, S., Murray, D.G., Steiner, B., Tucker, P.A., Vasudevan, V., Warden,
P., Wicke, M., Yu, Y., Zheng, X.: Tensorflow: {A} system for large-scale
machine learning. In: Keeton, K., Roscoe, T. (eds.) 12th {USENIX} Symposium
on Operating Systems Design and Implementation, {OSDI} 2016, Savannah, GA,
USA, November 2-4, 2016. pp. 265--283. {USENIX} Association (2016),
\url{https://www.usenix.org/conference/osdi16/technical-sessions/presentation/abadi}

\bibitem{avazu}
Avazu: Avazu click-through rate prediction (2015),
\url{https://www.kaggle.com/c/avazu-ctr-prediction/data}

\bibitem{criteo}
Criteo: Criteo display ad challenge (2014),
\url{http://labs.criteo.com/2014/02/kaggle-display-advertising-challenge-dataset}

\bibitem{adagrad}
Duchi, J., Hazan, E., Singer, Y.: Adaptive subgradient methods for online
learning and stochastic optimization. Journal of Machine Learning Research
\textbf{12},  2121--2159 (2011). \doi{10.5555/1953048.2021068},
\url{https://dl.acm.org/doi/10.5555/1953048.2021068}

\bibitem{auc}
Graepel, T., Candela, J.Q., Borchert, T., Herbrich, R.: Web-scale bayesian
click-through rate prediction for sponsored search advertising in microsoft's
bing search engine. In: F{\"{u}}rnkranz, J., Joachims, T. (eds.) Proceedings
of the 27th International Conference on Machine Learning (ICML-10), June
21-24, 2010, Haifa, Israel. pp. 13--20. Omnipress (2010),
\url{https://icml.cc/Conferences/2010/papers/901.pdf}

\bibitem{shampoo}
Gupta, V., Koren, T., Singer, Y.: Shampoo: Preconditioned stochastic tensor
optimization. In: Dy, J.G., Krause, A. (eds.) Proceedings of the 35th
International Conference on Machine Learning, {ICML} 2018,
Stockholmsm{\"{a}}ssan, Stockholm, Sweden, July 10-15, 2018. Proceedings of
Machine Learning Research, vol.~80, pp. 1837--1845. {PMLR} (2018),
\url{http://proceedings.mlr.press/v80/gupta18a.html}

\bibitem{ipinyou}
iPinYou: ipinyou global rtb bidding algorithm competition (2013),
\url{https://www.kaggle.com/lastsummer/ipinyou}

\bibitem{adam}
Kingma, D.P., Ba, J.L.: Adam: A method for stochastic optimization. In:
Proceedings of the 3rd International Conference on Learning Representations.
ICLR '15, San Diego, CA, USA (2015)

\bibitem{online_to_batch}
Littlestone, N.: From on-line to batch learning. In: Rivest, R.L., Haussler,
D., Warmuth, M.K. (eds.) Proceedings of the Second Annual Workshop on
Computational Learning Theory, {COLT} 1989, Santa Cruz, CA, USA, July 31 -
August 2, 1989. pp. 269--284. Morgan Kaufmann (1989),
\url{http://dl.acm.org/citation.cfm?id=93365}

\bibitem{ftrl}
McMahan, H.B.: Follow-the-regularized-leader and mirror descent: Equivalence
theorems and l1 regularization. In: Proceedings of the 14th International
Conference on Artificial Intelligence and Statistics. AISTATS '11, vol.~15,
pp. 525--533. PMLR, Fort Lauderdale, FL, USA (2011)

\bibitem{adclickprediction}
McMahan, H.B., Holt, G., Sculley, D., Young, M., Ebner, D., Grady, J., Nie, L.,
Phillips, T., Davydov, E., Golovin, D., Chikkerur, S., Liu, D., Wattenberg,
M., Hrafnkelsson, A.M., Boulos, T., Kubica, J.: Ad click prediction: a view
from the trenches. In: Proceedings of the 19th ACM SIGKDD international
conference on Knowledge discovery and data mining. pp. 1222--1230. KDD '13,
ACM, Chicago, Illinois, USA (2013)

\bibitem{ftrl2}
McMahan, H.B., Streeter, M.J.: Adaptive bound optimization for online convex
optimization. In: {COLT} 2010 - The 23rd Conference on Learning Theory,
Haifa, Israel, June 27-29, 2010. pp. 244--256. Omnipress (2010),
\url{http://colt2010.haifa.il.ibm.com/papers/COLT2010proceedings.pdf\#page=252}

\bibitem{dlrm}
Naumov, M., Mudigere, D., Shi, H.M., Huang, J., Sundaraman, N., Park, J., Wang,
X., Gupta, U., Wu, C., Azzolini, A.G., Dzhulgakov, D., Mallevich, A.,
Cherniavskii, I., Lu, Y., Krishnamoorthi, R., Yu, A., Kondratenko, V.,
Pereira, S., Chen, X., Chen, W., Rao, V., Jia, B., Xiong, L., Smelyanskiy,
M.: Deep learning recommendation model for personalization and recommendation
systems. CoRR  \textbf{abs/1906.00091} (2019),
\url{http://arxiv.org/abs/1906.00091}

\bibitem{primaldual2}
Nesterov, Y.E.: Smooth minimization of non-smooth functions. Math. Program.
\textbf{103},  127--152 (2005)

\bibitem{primaldual}
Nesterov, Y.E.: Primal-dual subgradient methods for convex problems. Math.
Program.  \textbf{120}(1),  221--259 (2009). \doi{10.1007/s10107-007-0149-x},
\url{https://doi.org/10.1007/s10107-007-0149-x}

\bibitem{groupftrl}
Ni, X., Yu, Y., Wu, P., Li, Y., Nie, S., Que, Q., Chen, C.: Feature selection
for facebook feed ranking system via a group-sparsity-regularized training
algorithm. In: Proceedings of the 28th ACM International Conference on
Information and Knowledge Management. pp. 2085--2088. CIKM '19, ACM, Beijing,
China (2019)

\bibitem{momentum2}
Polyak, B.T.: Some methods of speeding up the convergence of iteration methods.
USSR Computational Mathematics and Mathematical Physics  \textbf{4}(5),
1--17 (1964). \doi{10.1016/0041-5553(64)90137-5}

\bibitem{pnn}
Qu, Y., Cai, H., Ren, K., Zhang, W., Yu, Y., Wen, Y., Wang, J.: Product-based
neural networks for user response prediction. In: Bonchi, F.,
Domingo{-}Ferrer, J., Baeza{-}Yates, R., Zhou, Z., Wu, X. (eds.) {IEEE} 16th
International Conference on Data Mining, {ICDM} 2016, December 12-15, 2016,
Barcelona, Spain. pp. 1149--1154. {IEEE} Computer Society (2016).
\doi{10.1109/ICDM.2016.0151}, \url{https://doi.org/10.1109/ICDM.2016.0151}

\bibitem{amsgrad}
Reddi, S.J., Kale, S., Kumar, S.: On the convergence of adam and beyond. In:
Proceedings of the 6th International Conference on Learning Representations.
ICLR '18, OpenReview.net, Vancouver, BC, Canada (2018)

\bibitem{sgd}
Robbins, H., Monro, S.: A stochastic approximation method. The annals of
mathematical statistics pp. 400--407 (1951)

\bibitem{convex}
Rockafellar, R.T.: Convex Analysis. Princeton Landmarks in Mathematics and
Physics, Princeton University Press (1970)

\bibitem{groupsparse}
Scardapane, S., Comminiello, D., Hussain, A., Uncini, A.: Group sparse
regularization for deep neural networks. Neurocomputing  \textbf{241},
81--89 (2017). \doi{10.1016/j.neucom.2017.02.029},
\url{https://doi.org/10.1016/j.neucom.2017.02.029}

\bibitem{dcn}
Wang, R., Fu, B., Fu, G., Wang, M.: Deep {\&} cross network for ad click
predictions. In: Proceedings of the ADKDD'17, Halifax, NS, Canada, August 13
- 17, 2017. pp. 12:1--12:7. {ACM} (2017). \doi{10.1145/3124749.3124754},
\url{https://doi.org/10.1145/3124749.3124754}

\bibitem{dualavg}
Xiao, L.: Dual averaging method for regularized stochastic learning and online
optimization. Journal of Machine Learning Research  \textbf{11},  2543--2596
(2010). \doi{10.5555/1756006.1953017},
\url{https://dl.acm.org/doi/10.5555/1756006.1953017}

\bibitem{onlinegrouplasso}
Yang, H., Xu, Z., King, I., Lyu, M.R.: Online learning for group lasso. In:
Proceedings of the 27th International Conference on Machine Learning. pp.
1191--1198. ICML '10, Omnipress, Haifa, Israel (2010)

\bibitem{adahessian}
Yao, Z., Gholami, A., Shen, S., Keutzer, K., Mahoney, M.W.: {ADAHESSIAN:} an
adaptive second order optimizer for machine learning. CoRR
\textbf{abs/2006.00719} (2020), \url{https://arxiv.org/abs/2006.00719}

\bibitem{adadelta}
Zeiler, M.D.: Adadelta: An adaptive learning rate method. CoRR
\textbf{abs/1212.5701} (2012), \url{https://arxiv.org/abs/1212.5701}

\bibitem{prune}
Zhu, M., Gupta, S.: To prune, or not to prune: Exploring the efficacy of
pruning for model compression. In: 6th International Conference on Learning
Representations, {ICLR} 2018, Vancouver, BC, Canada, April 30 - May 3, 2018,
Workshop Track Proceedings. OpenReview.net (2018),
\url{https://openreview.net/forum?id=Sy1iIDkPM}

\end{thebibliography}
\end{document}